\documentclass{article} % For LaTeX2e
\usepackage{iclr2026_conference,times}

\usepackage{amsmath,amsfonts,bm}

\def\1{\bm{1}}

\def\vx{{\bm{x}}}

\def\mK{{\bm{K}}}

\def\mO{{\bm{O}}}
\def\mP{{\bm{P}}}
\def\mQ{{\bm{Q}}}
\def\mR{{\bm{R}}}
\def\mS{{\bm{S}}}

\def\mV{{\bm{V}}}

\DeclareMathAlphabet{\mathsfit}{\encodingdefault}{\sfdefault}{m}{sl}
\SetMathAlphabet{\mathsfit}{bold}{\encodingdefault}{\sfdefault}{bx}{n}

\def\gH{{\mathcal{H}}}

\def\gT{{\mathcal{T}}}

\def\gW{{\mathcal{W}}}

\newcommand{\E}{\mathbb{E}}

\newcommand{\R}{\mathbb{R}}

\usepackage{hyperref}
\usepackage{url}

\usepackage[table,xcdraw]{xcolor}
\usepackage{makecell}
\usepackage{booktabs}
\usepackage{multirow}
\usepackage{graphicx} 
\usepackage{subcaption}
\usepackage{amsthm} 
\newcommand{\eqby}[1]{\mathrel{\overset{\makebox[0pt]{\scriptsize #1}}{=}}}
\usepackage{algorithm}
\usepackage{algpseudocode}

\title{UltraViCo:~Breaking Extrapolation \\ Limits in Video Diffusion Transformers}

\author{Min Zhao$^{1,2}$~\thanks{$^\star$Equal contribution. $^\ddag$The corresponding author (\texttt{dcszj@tsinghua.edu.cn}).}~~, Hongzhou Zhu$^{1,2~*}$, Yingze Wang$^{1~*}$, Bokai Yan$^{3}$, Jintao Zhang$^{1,2}$, Guande \\ 
\textbf{He}$^{4}$, 
\textbf{Ling Yang}$^{5}$, \textbf{Chongxuan Li}$^{3}$, \textbf{Jun Zhu}$^{1,2 ^\ddag}$ \\
$^{1}$Dept. of Comp. Sci. \& Tech., BNRist Center, THU-Bosch ML Center, Tsinghua University. \\
$^{2}$ShengShu. $^{3}$Gaoling School of Artificial Intelligence, Renmin University of China. \\
$^{4}$The University of Texas at Austin. $^{5}$ Princeton University.\\
\texttt{gracezhao1997@gmail.com, zhuhz22@mails.tsinghua.edu.cn}
}
\newtheorem{proposition}{Proposition}

\iclrfinalcopy % Uncomment for camera-ready version, but NOT for submission.
\begin{document}

\maketitle

\begin{abstract}

Despite advances, video diffusion transformers still struggle to generalize beyond their training length, a challenge we term video length extrapolation. We identify two failure modes: model-specific \emph{periodic content repetition} and a universal \emph{quality degradation}.
Prior works attempt to solve repetition via positional encodings, overlooking quality degradation and achieving only limited extrapolation. In this paper, we revisit this challenge from a more fundamental view—attention maps, which directly govern how context influences outputs. We identify that both failure modes arise from a unified cause: \emph{attention dispersion}, where tokens beyond the training window dilute learned attention patterns. This leads to quality degradation and repetition emerges as a special case when this dispersion becomes structured into \emph{periodic attention patterns}, induced by harmonic properties of positional encodings. Building on this insight, we propose \emph{\textbf{UltraViCo}}, a training-free, plug-and-play method that suppresses attention for tokens beyond the training window via a constant decay factor. By jointly addressing both failure modes, we outperform a broad set of baselines largely across models and extrapolation ratios, pushing the extrapolation limit from $~2\times$ to $4\times$. Remarkably, it improves Dynamic Degree and Imaging Quality by 233\% and 40.5\% over the previous best method at $4\times$ extrapolation. Furthermore, our method generalizes seamlessly to downstream tasks such as controllable video synthesis and editing. Project page is available at \href{https://thu-ml.github.io/UltraViCo.github.io/}{https://thu-ml.github.io/UltraViCo.github.io/}.

\end{abstract}

\begin{figure}[h!]
  \centering

  \begin{subfigure}{0.93\textwidth}
    \centering
    
      \centering
      \includegraphics[width=\linewidth,keepaspectratio]{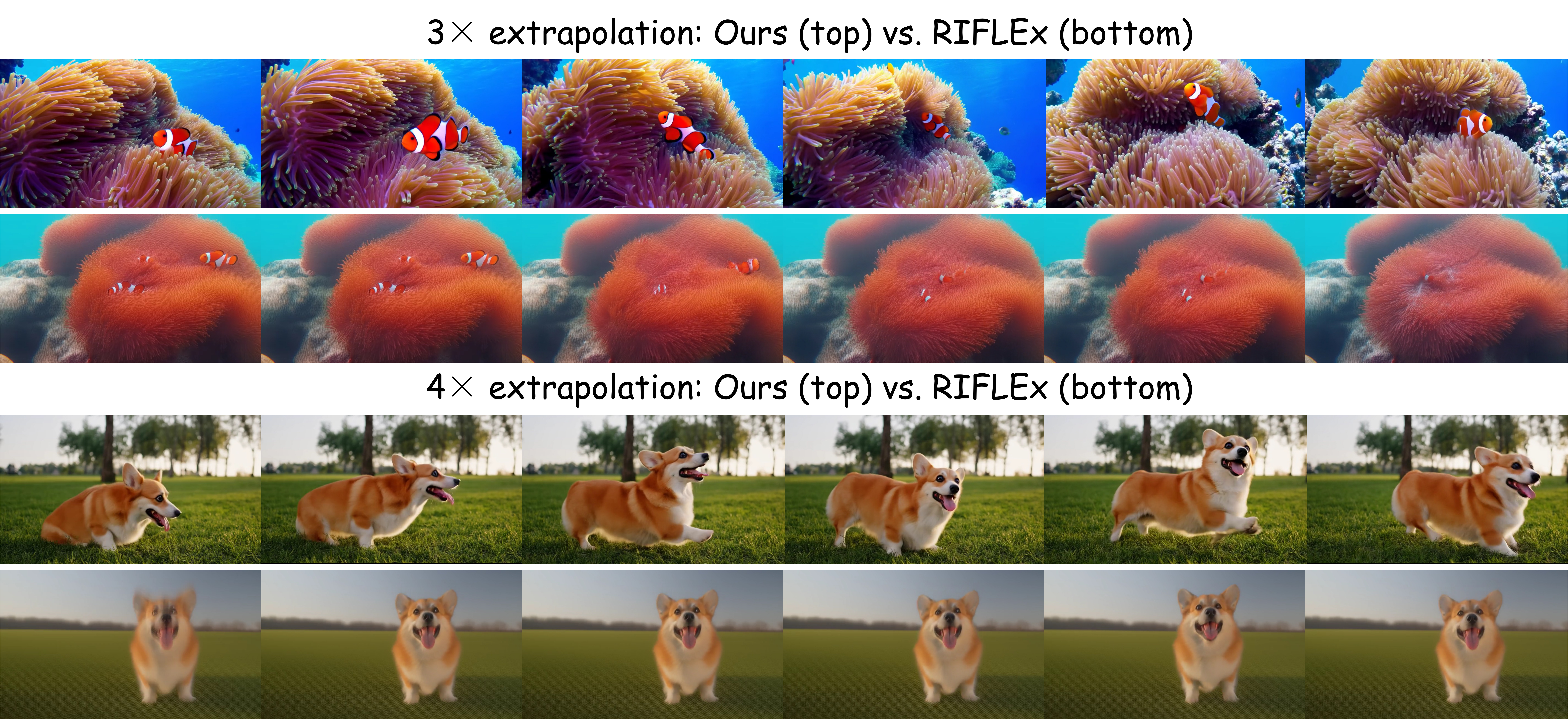}
    
    \subcaption{Extending T2V models up to $4\times$, where existing method yields nearly static, low-quality videos.}\label{fig:stack:a}
  \end{subfigure}

  \begin{subfigure}{0.93\textwidth}
    \centering
    
      \centering
      \includegraphics[width=\linewidth,keepaspectratio]{images/qualitive/downstream.pdf}
  
    \subcaption{Generalization to downstream tasks at $3\times$. See more tasks in Appendix~\ref{appendix: more_qualitive_experiments_wan_cog_3_4}.}\label{fig:stack:b}
  \end{subfigure}

  \caption{\textbf{Visual results.} UltraViCo achieves significant extrapolation improvement on (a) T2V models and (b) downstream tasks. \emph{See prompts and videos in supplementary materials.} }
  \label{fig:ability}
  \vspace{-0.5cm}
\end{figure}

% These observations raise three critical questions: (1) \textit{why does content repetition occur only in certain models?} (2) \textit{what is the cause of quality degradation?} (3) \textit{is there a unified mechanism underlying both failure modes?}

\section{Introduction}

% CogVideoX~\citep{yang2024cogvideox}, HunayunVideo~\citep{kong2024hunyuanvideo} and Wan~\citep{wan2025wan}

% Through extensive experiments with state-of-the-art (SOTA) models including CogVideoX~\citep{yang2024cogvideox}, HunayunVideo~\citep{kong2024hunyuanvideo} and Wan~\citep{wan2025wan}, we identify two failure modes:
% 5 s 从何而来。

Building upon the expressive power of diffusion transformers (DiTs)~\citep{bao2023all,peebles2023scalable}, recent advances in text-to-video (T2V) generation~\citep{bao2024vidu,zheng2024opensora,videoworldsimulators2024,wan2025wan,kong2024hunyuanvideo,hong2022cogvideo} have enabled models to synthesize high-fidelity videos. However, these models are typically trained on a fixed maximum sequence length (e.g., 5 seconds)~\citep{wan2025wan,kong2024hunyuanvideo,hong2022cogvideo} and struggle to generate videos beyond their training length, a task we term \emph{video length extrapolation}, which is critical for practical applications.

To investigate the core challenges of this task, we conduct experiments on a range of models and identify two failure modes: (i) a model-specific \emph{periodic content repetition}, where short clips 
loop indefinitely in certain models; and (ii) a universal \emph{quality degradation}, 
manifested as blurred spatial details and frozen temporal dynamics across all models. 
Both failures become increasingly severe as the extrapolation length grows.  Prior work, such as RIFLEx~\citep{zhao2025riflex}, tackles repetition from the perspective of positional encodings, while overlooking quality degradation and therefore achieving limited extrapolation. We contend, however, that positional encodings play only an \emph{indirect} role by perturbing queries and keys to influence attention. In contrast, attention itself—\emph{directly} aggregating contextual information to generate outputs—offers a more fundamental view.

Therefore, we revisit extrapolation failures through the lens of attention maps. Our systematic analysis of attention maps shows that both failure modes arise from a unified mechanism: \emph{attention dispersion}. This occurs when new tokens beyond the training length dilute the learned attention patterns. This leads to quality degradation and repetition arises as a special case when dispersion becomes organized into \emph{periodic attention patterns}. Specifically, this happens when positional encoding frequencies form \emph{harmonics}, enabling the largest-amplitude frequency and its harmonics to accumulate amplitude and contribute substantially to the overall amplitude.

Building on this unified view, we propose \emph{\textbf{Ultra}}-extrapolated \emph{\textbf{Vi}}deo via Attention \emph{\textbf{Co}}ncentration (\emph{\textbf{UltraViCo}}), a plug-and-play method that suppresses attention for tokens beyond the training window with a constant decay factor. This adjustment reallocates attention to reliable in-window context while naturally breaking periodic patterns, thus simultaneously addressing both failure modes. Notably, standard attention implementations encounter out-of-memory errors when modifying logits for long video sequences. We therefore develop a memory-efficient CUDA kernel that enables scalable applications on large video models.

To validate our approach, we conduct comprehensive evaluations on various T2V models~\citep{kong2024hunyuanvideo, yang2024cogvideox, wan2025wan} and extrapolation ratios, against a large family of baselines~\citep{chen2023extending,bloc97,zhuo2024lumina,peng2023yarn,zhao2025riflex}. Experiments demonstrate that our method consistently surpasses all baselines in all settings by simultaneously addressing both failure modes. Notably, while prior methods collapse beyond $3\times$ extrapolation and yield static videos, ours maintains fluid motion, effectively extending the practical limit from $2\times$ to $4\times$. Remarkably, it improves Dynamic Degree and Imaging Quality by 233\% and 40.5\% over the previous best method at $4\times$ extrapolation. Beyond this, our method also generalizes seamlessly to downstream tasks such as various controllable video synthesis and editing.

% same highlight points as abs

\begin{figure}[h!]
\centering
\resizebox{\textwidth}{!}{%
\begin{tabular}{c|c|c}
\toprule
\scriptsize \textbf{} & \small \textbf{HunyuanVideo} & \small \makecell{\textbf{Wan}}\\
\midrule
\multirow{2}{*}{\makecell[t]{\small \textbf{Normal}\\ \textbf{length}}}
&
\begin{minipage}{0.5\textwidth}\centering
\includegraphics[width=\textwidth]{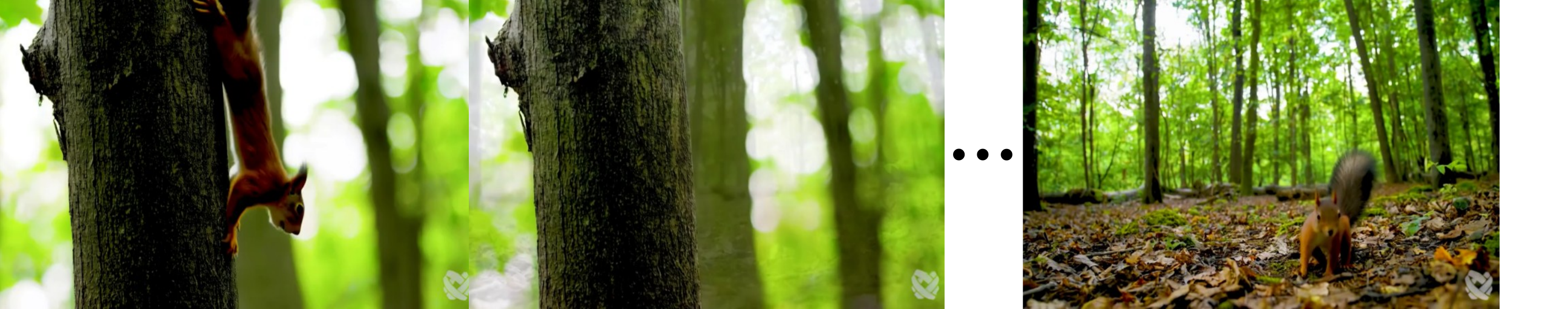}
\end{minipage}
&
\begin{minipage}{0.5\textwidth}\centering
\includegraphics[width=\textwidth]{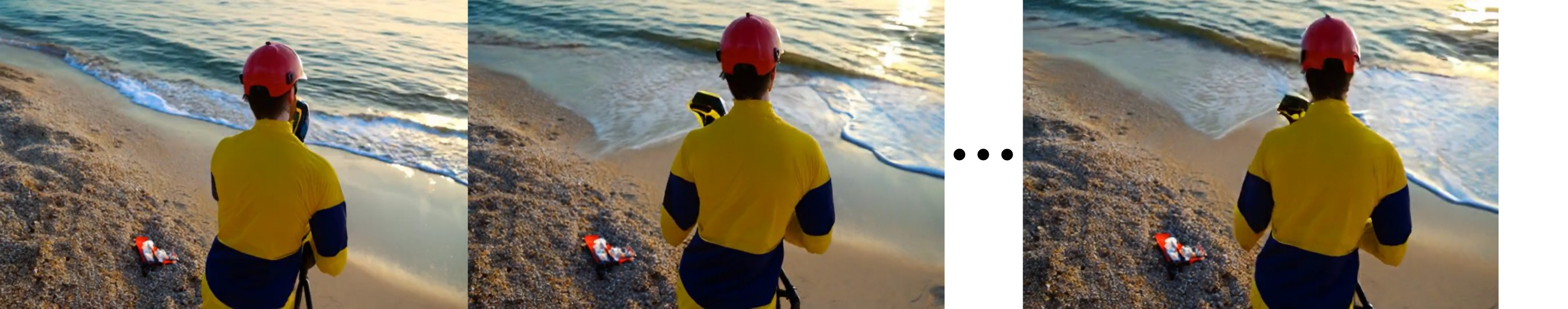}
\end{minipage}
\\
& \small{Video of $129$ frames} & \small{Video of $81$ frames}\\
\midrule

\multirow{2}{*}{\makecell[t]{\small \textbf{3 $\times$}\\ \textbf{extra.}}}
&
\begin{minipage}{0.5\textwidth}\centering
\includegraphics[width=\textwidth]{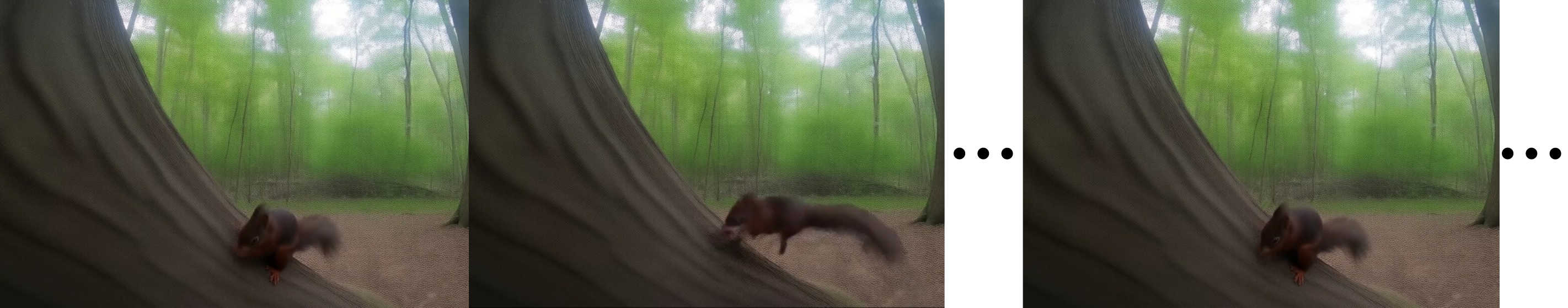}
\end{minipage}
&
\begin{minipage}{0.5\textwidth}\centering
\includegraphics[width=\textwidth]{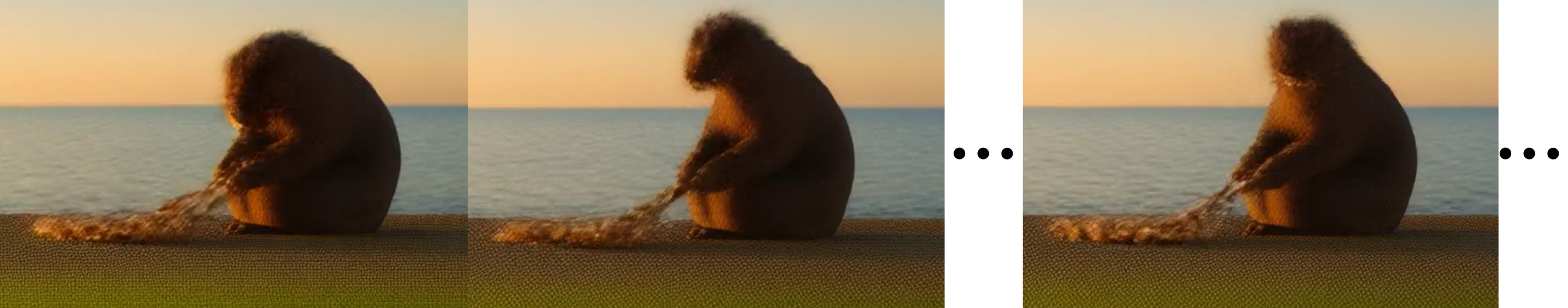}
\end{minipage}
\\
& \small{(a) Periodic content repetition and quality degradation.} &
  \small{(b) Quality degradation.}\\
\midrule

\multirow{2}{*}{\makecell[t]{\small \textbf{Variable}\\ \textbf{extra.}}}
& \multicolumn{2}{c}{%
  \begin{minipage}{\linewidth}\centering
  \includegraphics[width=\linewidth]{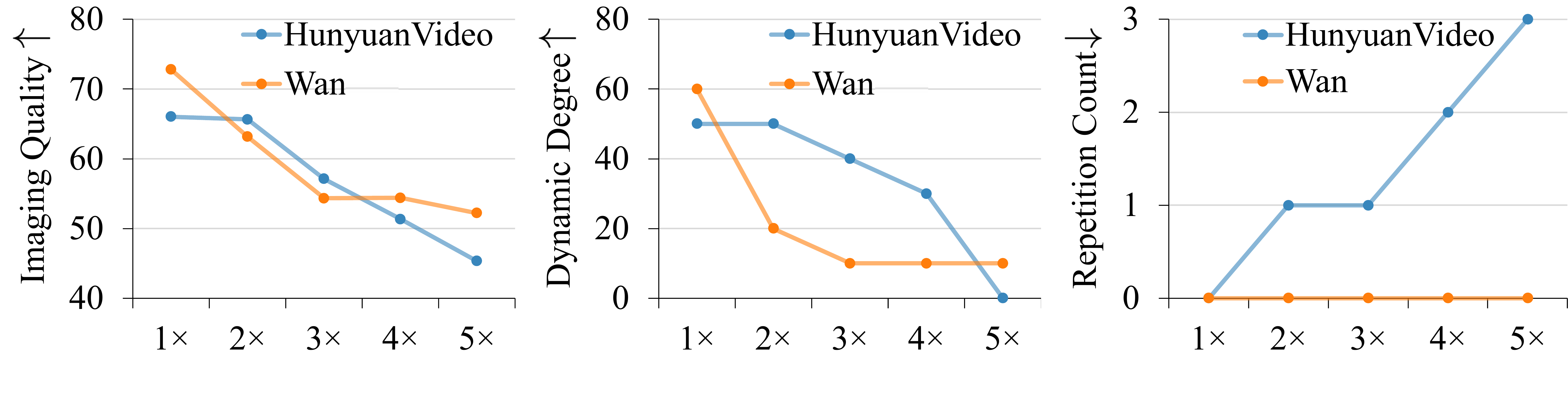}
  \end{minipage}
} \\
& \multicolumn{2}{c}{\small{(c) Both quality and repetition worsen as the extrapolation grows from $1\times$ to $5\times$.}}\\
\bottomrule
\end{tabular}%
}
\caption{\textbf{Failure modes of video length extrapolation.} Some models exhibit \emph{periodic content repetition}, while \emph{quality degradation} occurs universally. Both failure modes intensify with longer extrapolations. “extra.” denotes extrapolation. See Appendix~\ref{appendix:faliure-modes-of-CogVideoX} for additional models.}
\label{fig:challenge}
\vspace{-0.5cm}
\end{figure}

\section{Preliminary}
\label{sec:rope}
\textbf{Attention mechanism with rotary position embedding.} Modern video diffusion models are largely built on DiTs whose core is the attention mechanism~\citep{vaswani2017attention, li2025radial}. The input video is patched into $L$ tokens, each projected into queries, keys, and values. To encode the position information, DiTs mainly adopt Rotary Position Embedding (RoPE)~\citep{su2024roformer}, which injects position into queries and keys through complex rotations. Concretely, for each query or key vector $\vx \in \mathbb{R}^D$ at position $t$, RoPE maps it to $\mathbb{R}^D$ as
\begin{align}
\label{eq:RoPE}
\boldsymbol{f}^{\text{RoPE}}(\vx, t)_i
= R_i(t) 
\begin{bmatrix}
   x_{2i} \\
   x_{2i+1}
\end{bmatrix}, \,
R_i(t) =
\begin{bmatrix}
   \cos (\phi_i t) & -\sin  (\phi_i t)\\
   \sin  (\phi_i t)& \cos  (\phi_i t)
\end{bmatrix},\,
i\in\{0,\dots,D/2-1\}.
\end{align}
Here, each frequency $\phi_i$ depends exponentially on $i$ and is used to encode the $(2i,2i+1)$ components of $\vx$. After RoPE, the queries and keys form matrices $\mQ \in \mathbb{R}^{L \times D}$ and $\mK \in \mathbb{R}^{L \times D}$. Their interaction yields the attention logits $\mS \in \mathbb{R}^{L \times L} $, which are normalized by the softmax function to obtain the attention scores $\mP \in \mathbb{R}^{L \times L}$. These scores are then applied to the value matrix $\mV \in \mathbb{R}^{L \times D'}$ to produce the output $\mO \in \mathbb{R}^{L \times D'}$:
\begin{align}
\label{eq:attention}
    \mS=\mQ\mK^\top, \quad\mP=\text{softmax}(\frac{\mS}{\sqrt{D}}), \quad\mO=\mP\mV.
\end{align}
For videos with temporal and spatial axes, Multimodal RoPE (M-RoPE)~\citep{wang2024qwen2} partitions the dimension $D=d_\gT+d_\gH+d_\gW$ and encodes each subspace separately. Since we focus on temporal extrapolation, we consider only the temporal axis and denote $d_\gT$ as $d$ for simplicity~(see details in Appendix~\ref{appendix: M-RoPE}). 

\textbf{Problem setting: video length extrapolation.} Despite advances, DiT-based video generation models struggle to produce videos longer than their training duration. This task, known as \emph{video length extrapolation }~\citep{zhao2025riflex}, aims to adapt a pre-trained model to generate high-quality videos of a sequence length $L'$ that exceeds its training length $L$, with the extrapolation ratio defined as $s = L'/L > 1$. Notably, video length extrapolation targets the model’s intrinsic ability to generate longer sequences in a single forward generation, which is orthogonal to prior methods~\citep{qiu2023freenoise,wang2023gen,kim2024fifo,wang2024loong,lu2024freelong} that rely on inference-time modifications. See Appendix~\ref{sec: related work} for more related work.

\section{Method}

\subsection{Failure Modes of Video Length Extrapolation}
\label{sec: failure modes}

In this section, we investigate the core challenges of video length extrapolation on a range of SOTA video diffusion transformers, including Wan~\citep{wan2025wan}, HunyuanVideo~\citep{kong2024hunyuanvideo}, and CogVideoX~\citep{yang2024cogvideox}.

Qualitative results in Fig.\ref{fig:challenge}a and Fig.\ref{fig:challenge}b reveal two distinct failure modes. The first is a \emph{periodic content repetition}, which occurs in certain models such as HunyuanVideo and CogVideoX. The second is a universal \emph{quality degradation}, characterized by compromised spatial fidelity and temporal dynamics across all models. To further investigate their trends across extrapolation lengths, we perform a quantitative analysis on 10 prompts using metrics including Imaging Quality~\citep{huang2024vbench}, Dynamic Degree~\citep{huang2024vbench}, and Repetition Count. Fig.~\ref{fig:challenge}c confirms that both failures become more severe as the extrapolation factor increases.

These findings raise three critical questions: First, \textit{why does periodic content repetition only manifest in specific models?} Second, \textit{what is the underlying cause of the universal quality degradation?} Most importantly, \textit{is there a unified cause behind these two seemingly independent failure modes?}

Existing work such as RIFLEx addresses only content repetition, neglecting quality degradation, which limits both model generalization and extrapolation capacity. While RIFLEx attributes repetition to positional encoding periodicity, we argue that positional encodings play only an indirect role by modulating queries and keys. Instead, as Eq.~\eqref{eq:attention} shows, the attention map itself is fundamental, since it directly determines how context is aggregated. This motivates us to revisit extrapolation failures through attention analysis.

\subsection{Attention Analysis of the Cause}
\label{sec:root_cause}

In this section, we first focus on the specific issue of periodic content repetition (Sec.~\ref{sec:repetition_mechanism}). Through an in-depth attention analysis of its underlying mechanism, we find, surprisingly, that the solution designed to resolve repetition also improves video quality. This key finding then allows us to understand the cause of the more universal problem of quality degradation (Sec.~\ref{sec: dispersion}), and ultimately reveals the intrinsic connection between the two failure modes.

\subsubsection{The Cause of Content Repetition: Periodic Attention Patterns}
\label{sec:repetition_mechanism}

\begin{figure}[h!]
    \centering
    \resizebox{\textwidth}{!}{
    \begin{tabular}{c|c|c}
    \toprule
    \scriptsize \small \textbf{Model} &\small \makecell{\textbf{Attention maps} } & 
    \small \makecell{\textbf{Statistical row-wise attention analysis}}\\ \midrule 
\multirow{2}{*}{\makecell[t]{\small \textbf{Hun.}}} &
    
    \begin{minipage}{0.2\textwidth}
    \centering
\includegraphics[width=\textwidth]{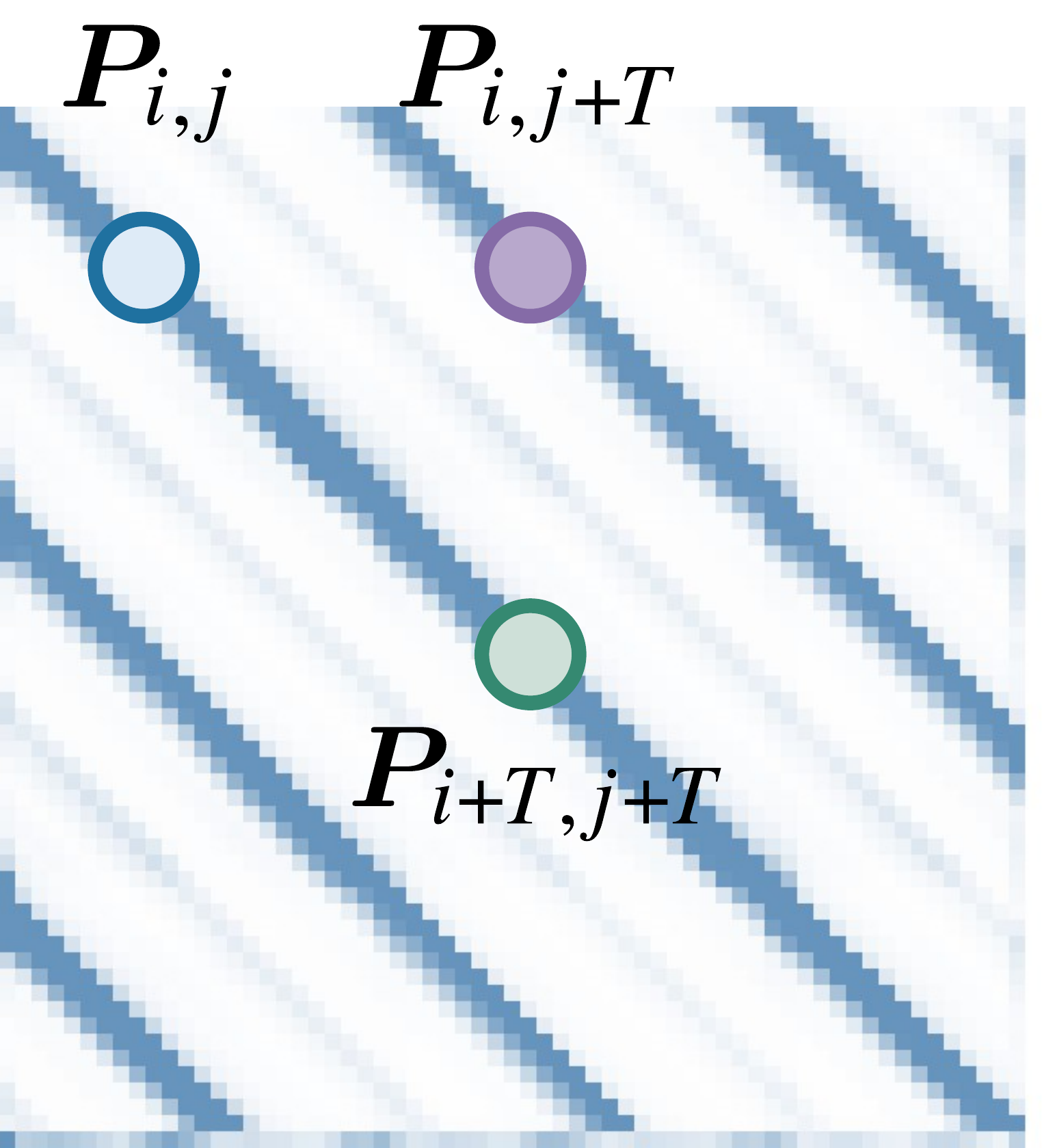}

    \end{minipage}&
    \begin{minipage}{0.8\textwidth}
    \centering   
    \includegraphics[width=\textwidth]{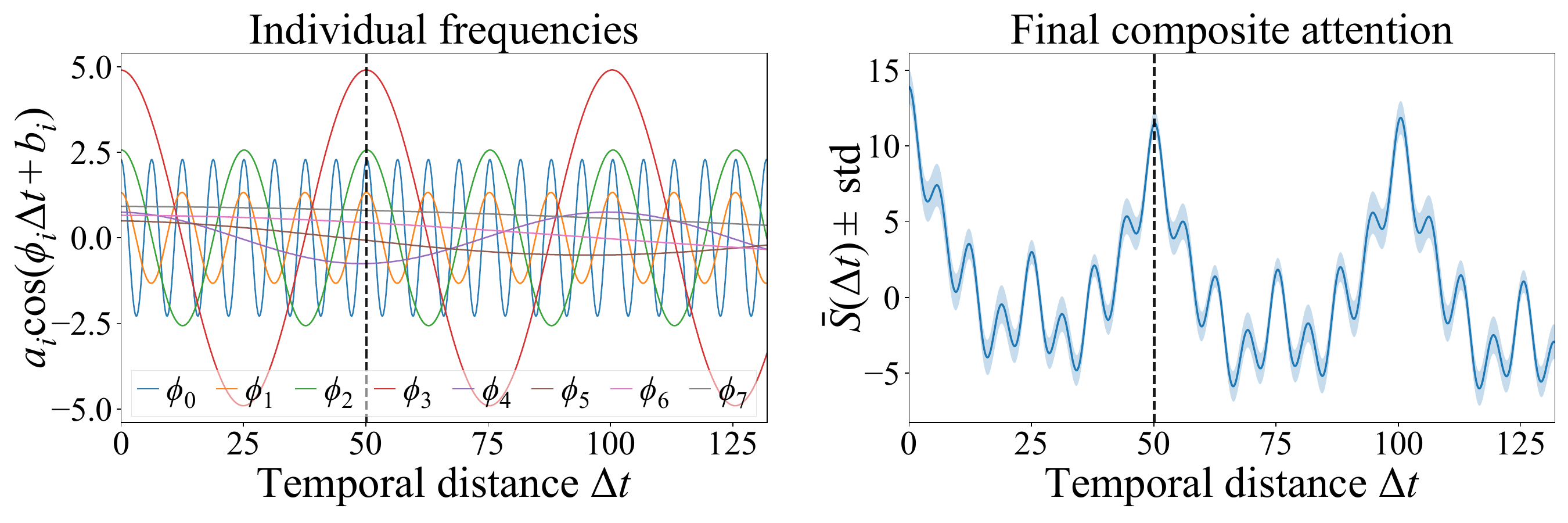}
    
    \end{minipage}
    \\
    &\small{(a) Periodic attention:}
    & \small{(b) Harmonic RoPE frequencies ($\phi_i/\phi_{N-1}\in \mathbb{N}^+$) amplify the largest-amplitude
    }\\
    &\small{$\mP_{i,j}\approx\mP_{i,j+T}$}
    & \small{frequency and its harmonics (dashed line), inducing periodic composite attention.}\\
    \midrule
\multirow{2}{*}{\makecell[t]{\small \textbf{Wan}}} &

    \begin{minipage}{0.2\textwidth}
    \centering
\includegraphics[width=\textwidth]{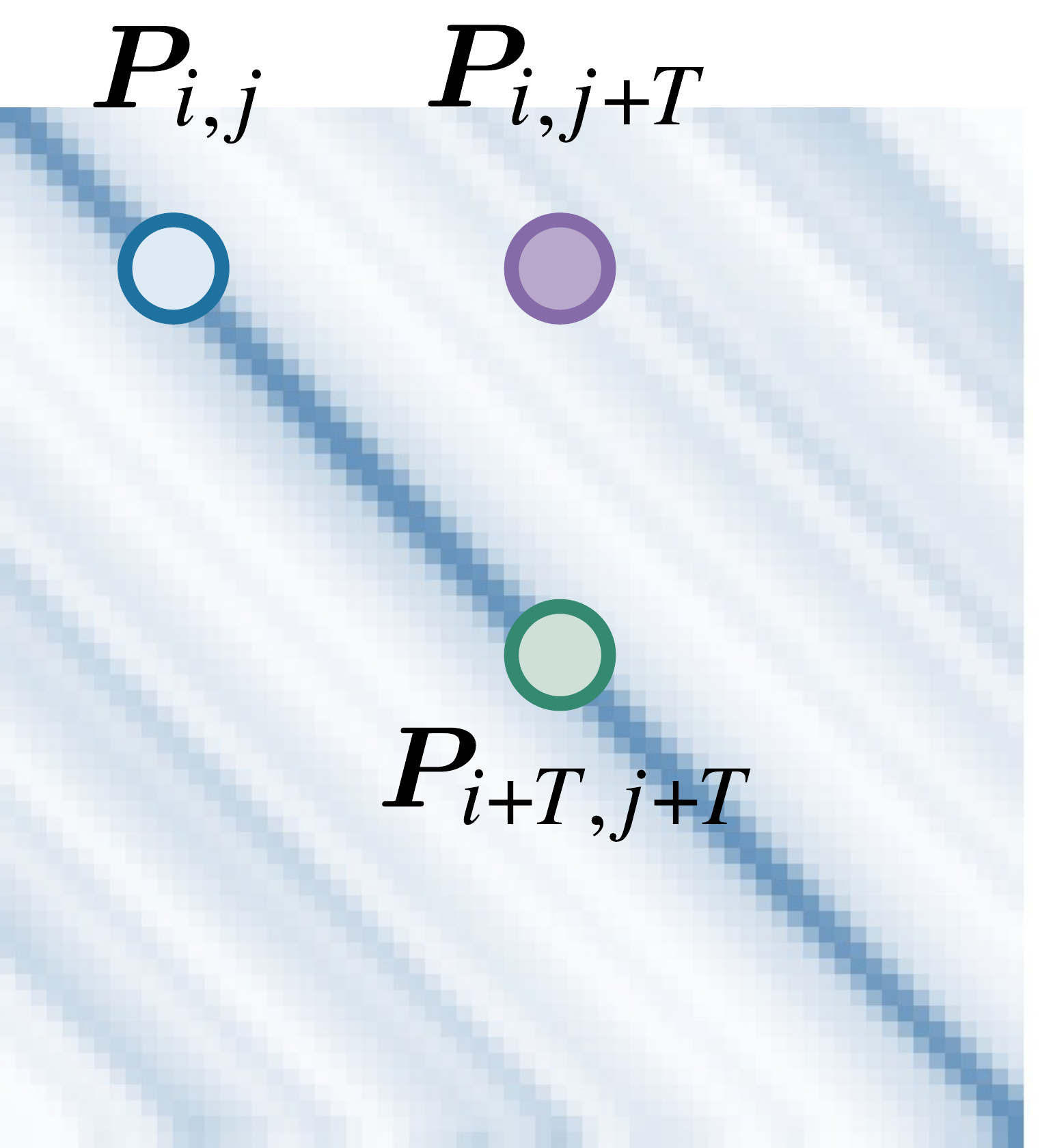}

    \end{minipage}
    & 
    \begin{minipage}{0.8\textwidth}
    \centering   
    \includegraphics[width=\textwidth]{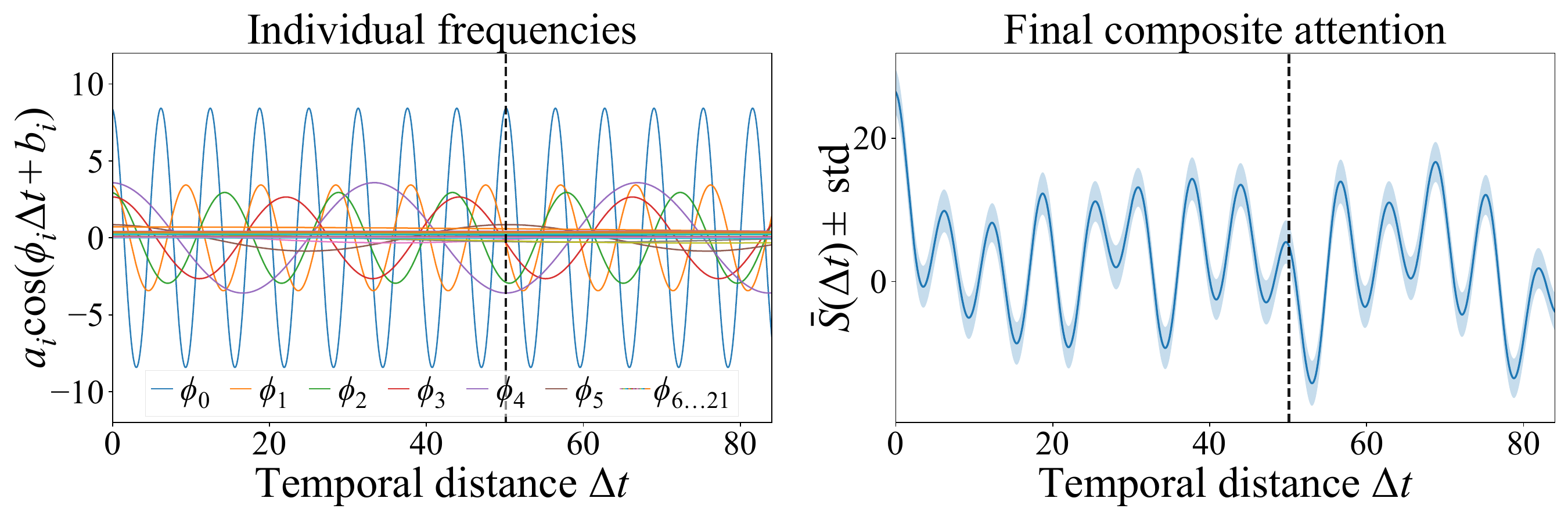}

    \end{minipage}
    \\ 
    &\small{(c) Non-periodic attention:}
     & \small{(d) Inharmonic RoPE frequencies ($\phi_i/\phi_{N-1}\notin \mathbb{N}^+$) disperse spectrum (dashed}
      \\ 
    &\small{
    $\mP_{i,j}\ne \mP_{i,j+T}$}
     & line), yielding non-periodicity in the final composite attention.
     
     % where no single frequency dominates,
     
     % yielding non-periodicity in the final composite attention.
     
     % where no single frequency dominates, yielding non-periodicity.
     
     \\
     \bottomrule
    \end{tabular}
    }
    \caption{\textbf{Periodic attention patterns as cause of content repetition.} Left: unlike Wan, HunyuanVideo exhibits row-wise periodic attention during $4\times$ extrapolation, causing repeated outputs. 
    Right: statistical row-wise attention can be expressed as a linear combination of trigonometric functions of RoPE frequencies, whose properties govern this periodicity. Hun. denotes HunyuanVideo.}
    \label{fig:repeat}
    \vspace{-0.5cm}
\end{figure}

\textbf{Periodic attention induces output repetition.}
We analyze the cause of content repetition by inspecting the attention map $\mP \in \mathbb{R}^{L' \times L'}$ during $4\times$ extrapolation, where $L'$ is the extrapolated sequence length (i.e., video features flattened into a 1D sequence). The entry at row $i$, column $j$ of $\mP$, denoted $P_{ij}$, is the attention score from query $i$ to key $j$. As shown in Fig.~\ref{fig:repeat}a, the attention map of HunyuanVideo reveals two properties that jointly induce periodic outputs.

% \begin{figure}[t]
%     \centering
%     \begin{subfigure}[t]{\dimexpr\textwidth*20/100\relax}
%         \centering
%         \includegraphics[height=0.27\textheight,keepaspectratio]{images/repetitive_map/attn_map.pdf}
%         \subcaption{caption(a)}\label{fig:top-p-dropout:quant}
%     \end{subfigure}\hfill
%     \begin{subfigure}[t]{\dimexpr\textwidth*80/100\relax}
%         \raggedleft
%         \includegraphics[height=0.27\textheight,keepaspectratio]{images/repetitive_map/statistics.pdf}
%         \subcaption{caption(b)}\label{fig:top-p-dropout:qual}
%     \end{subfigure}
% \vspace{-.25cm}
%     \caption{
%     \textbf{Cause of content repetition: periodic attention patterns.} Left: Unlike Wan, HunyuanVideo exhibits periodic attention patterns during extrapolation, leading to periodicity repetition in outputs. Right: .}
%     \label{fig:top-p-dropout}
% \end{figure}

\renewcommand\cellset{\renewcommand\arraystretch{0.7}}

First, the map exhibits a distinct \emph{row-wise periodicity}. Specifically, for any query at position $i$, its attention scores to key positions $j$ and $j+T$ are nearly identical: $\mP_{i,j} \approx \mP_{i,j+T}$, where $T$ corresponds to the observed repetition period in Sec.~\ref{sec: failure modes}. As indicated in Fig.~\ref{fig:repeat}a, the blue and purple circles highlight nearly equal scores.  Second, the map shows \emph{relative positional invariance}: query–key pairs with the same relative displacement $p$ yield approximately equal scores, $\mP_{i,j}\approx \mP_{i+p,j+p}$. This RoPE-induced property appears as uniform values along diagonals and subdiagonals; for example, when $p=T$, the scores marked by the blue and green circles are nearly identical.

Combining these properties, we can derive that entire query rows also repeat periodically: $\mP_{i+T,j} \approx \mP_{i,j}$, as shown by the green and purple circles. Thus, rows $i$ and $i+T$ retrieve nearly the same weighted information from the value $\mV$, leading to periodic outputs (see Appendix~\ref{appendix: derivation of the periodic outputs} for details):
\begin{equation}
\label{eq: periodic attention}
\mO_{i+T} = \sum_{j=0}^{L'-1} \mP_{i+T,j}\mV_j \approx \sum_{j=0}^{L'-1} \mP_{i,j}\mV_j = \mO_i.
\end{equation}
This periodicity is directly reflected in repeated content in pixel space. Larger extrapolation ratios traverse more periods, thus increasing repetition counts, which is consistent with our observations in Sec.~\ref{sec: failure modes}. By contrast, the attention map of Wan (Fig.~\ref{fig:repeat}c) does not display such row-wise periodicity, and accordingly its outputs remain free of repetition.

\textbf{Origin of periodic attention patterns.} Next, we show that such model-specific row-wise periodicity originates from the RoPE frequencies. To reveal the core row-wise attention structure from noise, we construct a statistical row attention pattern $\bm{\bar{S}}(\Delta t)$, which captures the relation between a query and keys at the same spatial location but $\Delta t$ latent frames apart. This is achieved by taking the expectation of the pre-softmax attention logits across all layers, heads, and query positions. As derived in Appendix~\ref{appendix: derivation of expectation} (based on Eq. (\ref{eq:attention})), this quantity admits the following trigonometric decomposition:
\begin{align}
\label{eq:trig_decomposition}
\bm{\bar{S}}(\Delta t) 
= \sum_{i=0}^{d/2-1} a_i \cos(\phi_i \Delta t + b_i) + C,
\end{align}
where $\{\phi_i\}_{i=0}^{d/2-1}$ are the RoPE frequencies defined in Sec.~\ref{sec:rope}, and $\{a_i\}_{i=0}^{d/2-1}, \{b_i\}_{i=0}^{d/2-1}, C$ are constants determined by the statistics of queries and keys from models, with $b_i$ typically close to zero. Visualizations of these frequency components for HunyuanVideo and Wan highlight a crucial difference (Fig.~\ref{fig:repeat}b,d, left). The periodicity of such a superposition is decided by the frequency relationships, as formalized in Proposition~\ref{prop:periodicity}.

\begin{proposition}[Period and Amplitude of Harmonics]
\label{prop:periodicity}
For a function $f(\Delta t) = \sum_{i=0}^{N-1} a_i \cos(\phi_i \Delta t)$, where $a_i > 0, \phi_i > 0$ and $\min_{i}\phi_i=\phi_{N-1}$, if and only if \, $ \forall i,\ \phi_i/\phi_{N-1}\in\mathbb{N}^+$ (i.e., they form a set of \textbf{harmonics}), $f(\Delta t)$ is periodic with period $T_{N-1}=\frac{2\pi}{\phi_{N-1}}$. In this case,
$\max_{\Delta t} f(\Delta t) = \sum_{i=0}^{N-1} a_i,
$
whenever $\Delta t=mT_{N-1},\,m\in\mathbb{Z}$ (i.e., whenever $\Delta t$ is at \textbf{harmonic alignment positions}).
\end{proposition}

We find that HunyuanVideo’s frequencies satisfy this \emph{harmonic} condition in Proposition~\ref{prop:periodicity}, allowing amplitude accumulation of the largest-amplitude frequency $\phi_3$ and its harmonics ($i < 3$) at \emph{harmonic alignment positions} $mT$ (dashed line in Fig.~\ref{fig:repeat}b), where $m \in \mathbb{Z}$. This yields a dominant component that contributes 79.6\% of the total amplitude, producing a strongly periodic composite attention pattern (Fig.~\ref{fig:repeat}b, right). A similar harmonic alignment is also observed in CogVideoX (Appendix~\ref{appendix: lemma remarks}). In contrast, Wan’s frequencies are not harmonically aligned, resulting in a dispersed spectrum where no frequency dominates (largest 31.6\%), and thus no clear periodicity emerges (Fig.~\ref{fig:repeat}d). Notably, while the strict periodicity of HunyuanVideo is determined by the lowest frequency, its small amplitude and long period make it negligible; the observed periodicity $T$ is effectively governed by the dominant frequency (see Appendix~\ref{appendix: lemma remarks}).

\begin{figure}[h!]
    \centering
    \resizebox{\textwidth}{!}{
    \begin{tabular}{c|c|c}
    \toprule
    \scriptsize \small \textbf{Model} &\small \makecell[t]{\textbf{Generated videos: baseline vs. intervention}} & \small \makecell[t]{\textbf{Attention maps: baseline vs. intervention}}\\ \midrule 

\multirow{2}{*}{\makecell[t]{\small \textbf{Hun.}}} &
    
    \begin{minipage}{0.61\textwidth}
    \centering
\includegraphics[width=\textwidth]{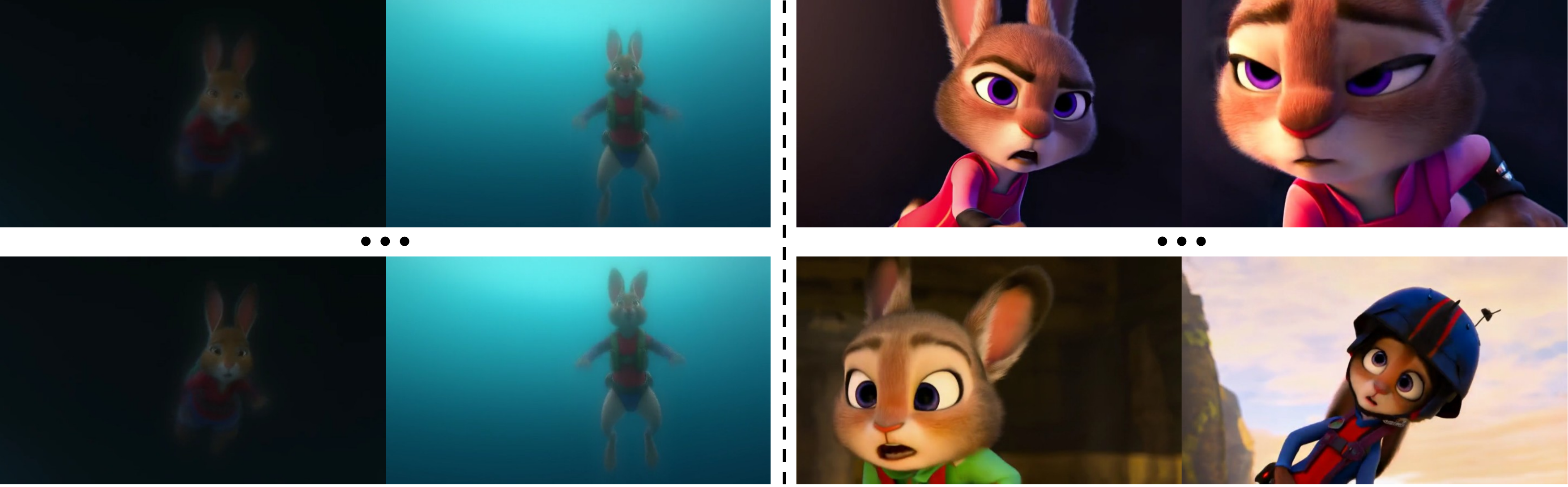}
% \vspace{.005cm}
    \end{minipage}&
    \begin{minipage}{0.39\textwidth}
    \centering
\includegraphics[width=\textwidth]{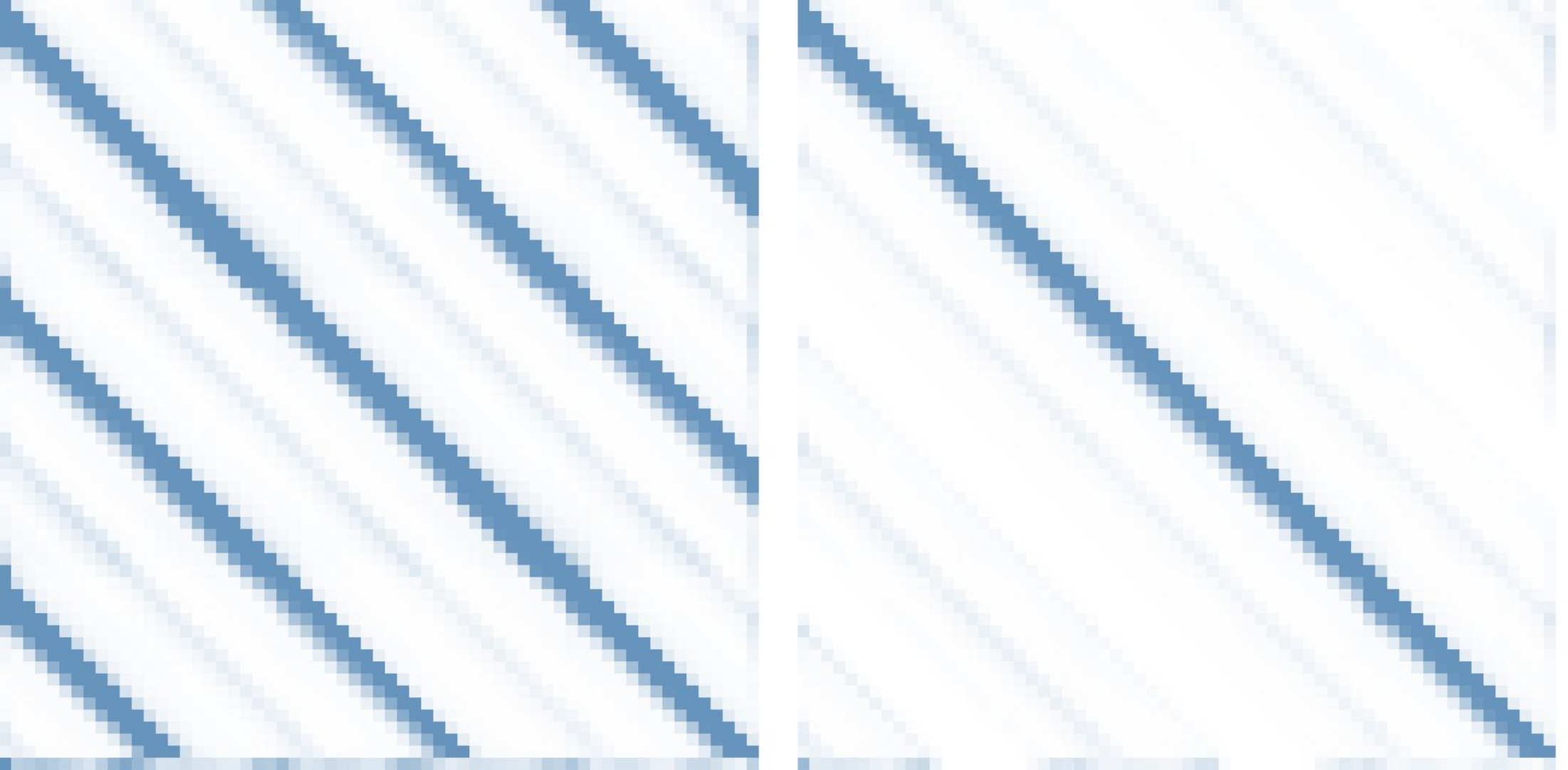}
 % \vspace{.005cm}
    \end{minipage}
    \\
    &\small{(a) Non-repetition and improved video quality  after intervention}
    & \small{(b) Attention focused centrally after intervention}\\
    \midrule

\multirow{2}{*}{\makecell[t]{\small \textbf{Wan}}} &

    \begin{minipage}{0.61\textwidth}
    \centering
\includegraphics[width=\textwidth]{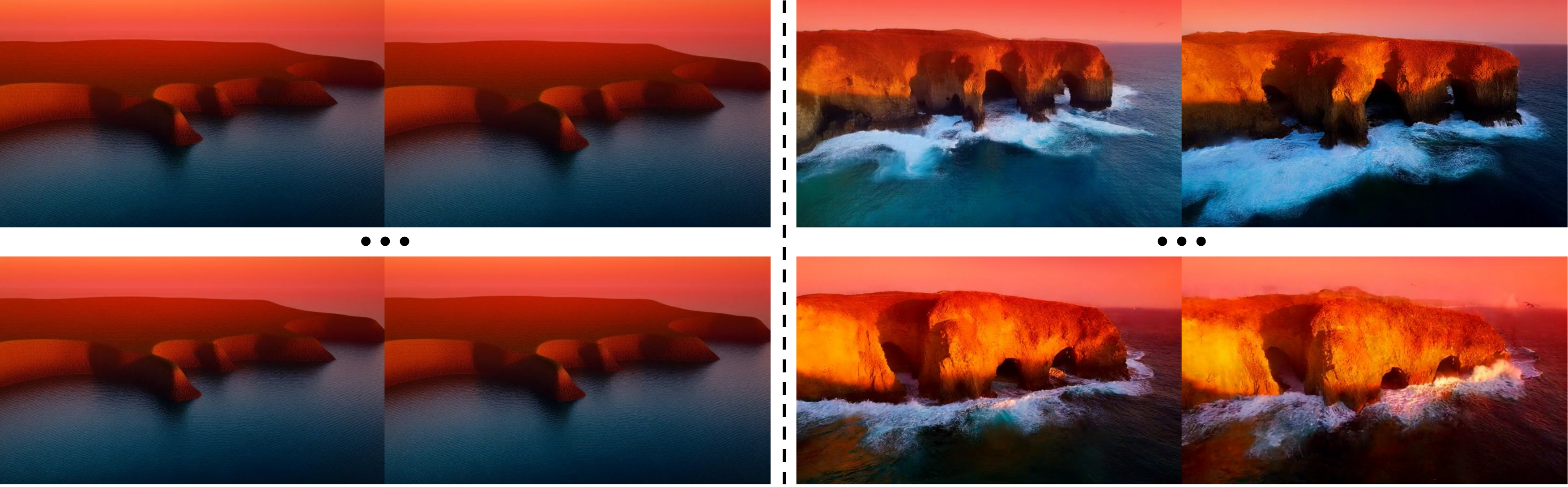}
% \vspace{.005cm}
    \end{minipage}&
    \begin{minipage}{0.39\textwidth}
    \centering
    \includegraphics[width=\textwidth]{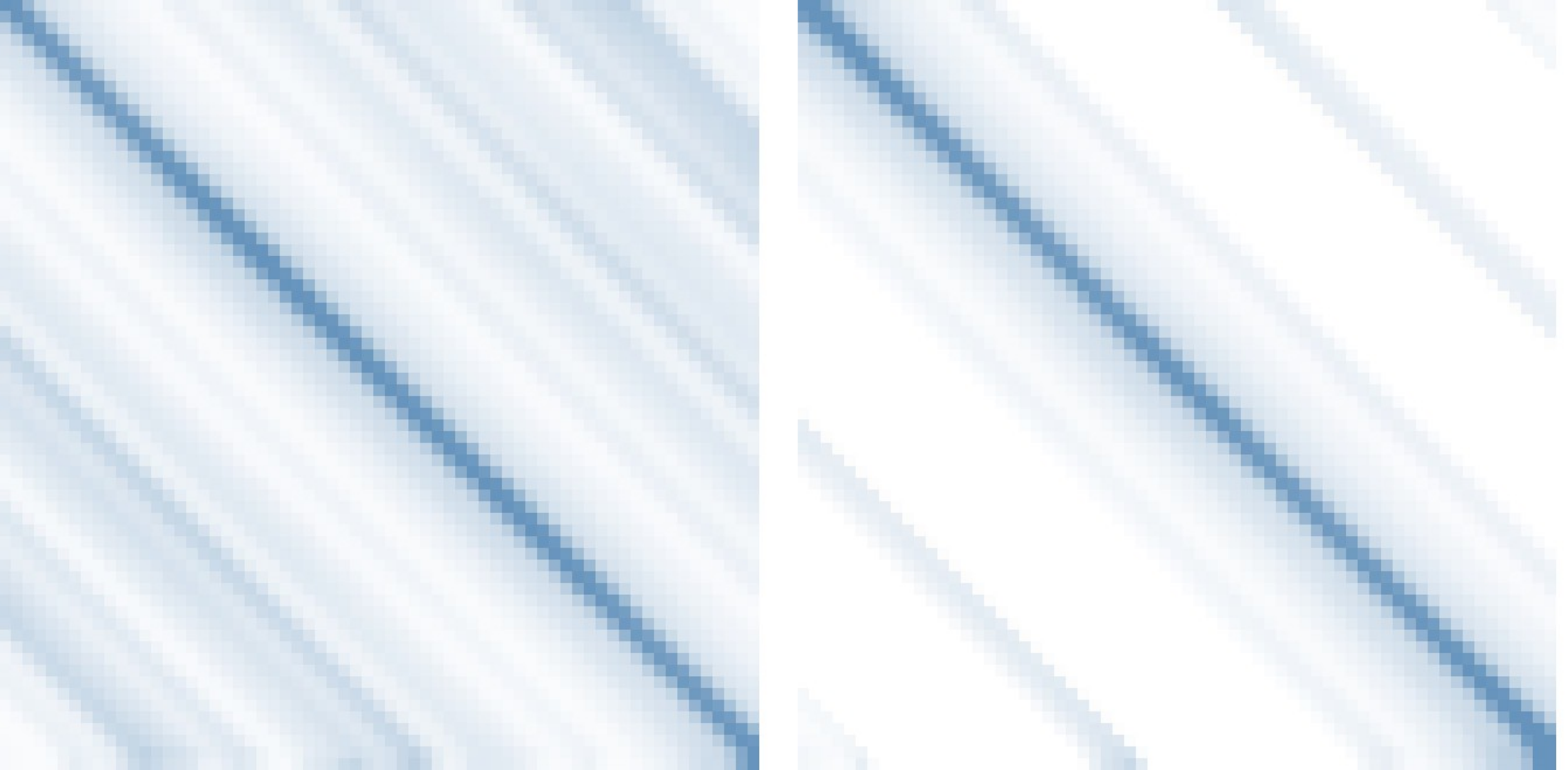}
    % \vspace{.005cm}
    \end{minipage}
    \\ 
    &\small{(c) Improved video quality after intervention}
    & \small{(d) Attention focused centrally after intervention}\\
     \bottomrule
    \end{tabular}
    }
    \caption{\textbf{Fixing repetition reveals attention dispersion as the fundamental cause.} 
    Left: our intervention, initially targeting repetition, surprisingly enhances video quality in both models. Right: the shared mechanism is revealed, where the intervention refocuses diffuse baseline attention toward the central training window. This suggests attention dispersion as the unified cause. 
    }
    \label{fig: dispersion}
\vspace{-0.5cm}
\end{figure}

In summary, our analysis establishes the causal chain: \textit{RoPE-induced frequency harmonics lead to periodic attention patterns, which in turn produce periodic output features and ultimately manifest as content repetition}. To validate this, we mask tokens at harmonic alignment positions $mT$. Breaking these constructive interference points disrupts periodic attention and, as shown in Fig.~\ref{fig: dispersion}a, effectively mitigates repetition.

\renewcommand\cellset{\renewcommand\arraystretch{0.7}}

\subsubsection{The Cause of Quality Degradation: Attention Dispersion}
\label{sec: dispersion}

Surprisingly, we find the above repetition-resolving intervention also improves video quality across both models (Fig.~\ref{fig: dispersion}a, c). This finding suggests a more profound hypothesis: content repetition and quality degradation may arise from a shared, fundamental underlying mechanism.

A comparison of attention maps shows our intervention consistently concentrates the initially diffuse attention (Fig.~\ref{fig: dispersion}b, d). This occurs because masking the harmonic peaks forces a softmax re-normalization, which sharpens the attention distribution by proportionally increasing the remaining scores. To further identify where this sharpened focus is most beneficial, we systematically masked different attention regions and found that concentrating attention within the original central training window yielded the strongest improvements (see details in Appendix~\ref{appendix: training window}). This leads us to hypothesize that \emph{attention dispersion} is the underlying issue. 
New tokens during extrapolation dilute the learned attention patterns within the original training window. This dispersion has two detrimental effects. Spatially, the model needs to consider far-away extrapolated frames, which makes it difficult to focus on fine details and results in visual blurriness. Temporally, taking these distant frames into account mixes local motion with unrelated movements, causing the video to appear static and unnatural. These effects are consistent with the quality degradation observed in Sec. ~\ref{sec: failure modes}.

\begin{figure}[h!]
    \centering
    \begin{subfigure}{\dimexpr\textwidth*35/100\relax}
        \centering
        \includegraphics[height=0.26\textheight,keepaspectratio]{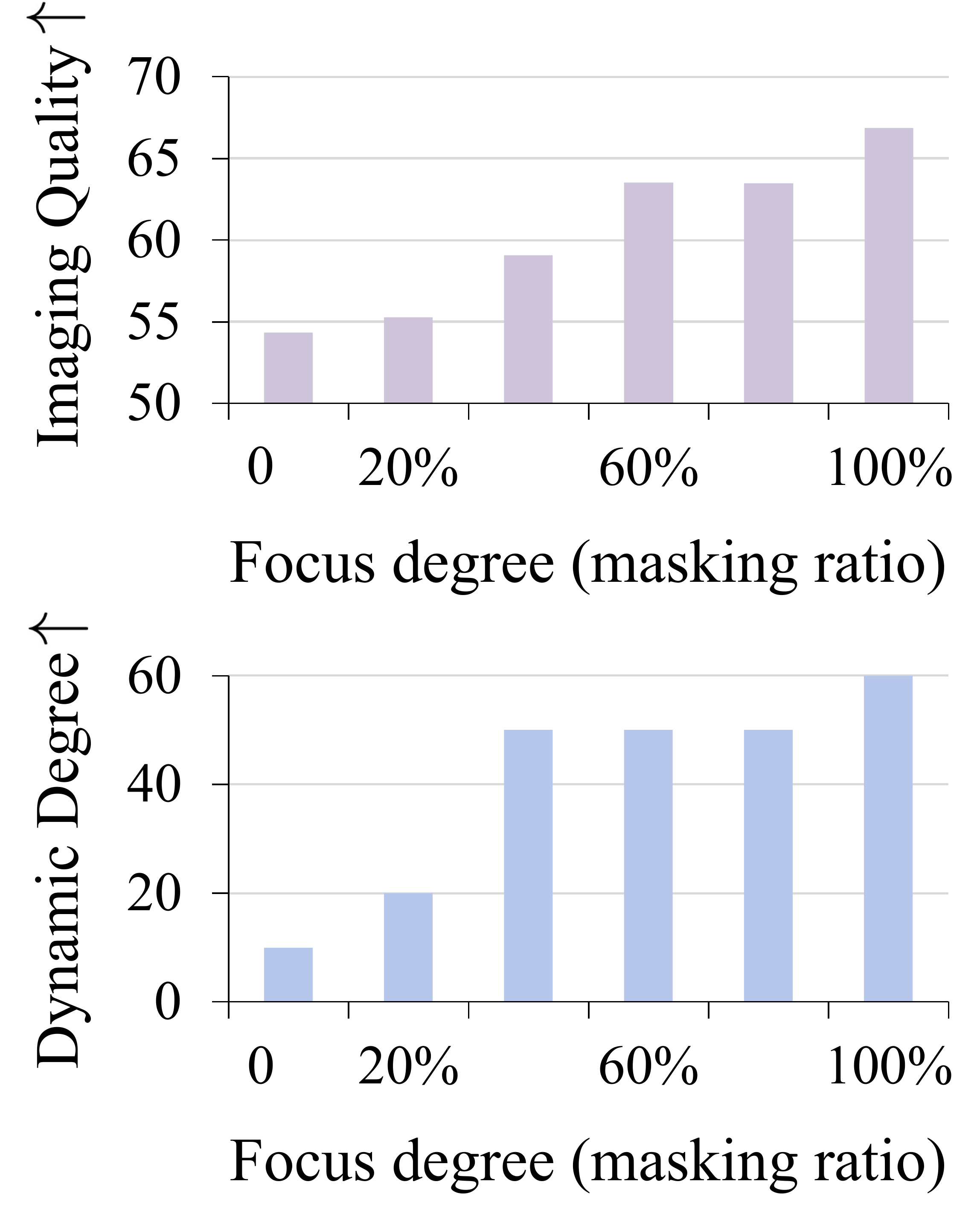}
        \subcaption{Quantitative results.}\label{fig:top-p-dropout:quant}
    \end{subfigure}\hfill
    \begin{subfigure}{\dimexpr\textwidth*65/100\relax}
        \raggedleft
        \includegraphics[height=0.26\textheight,keepaspectratio]{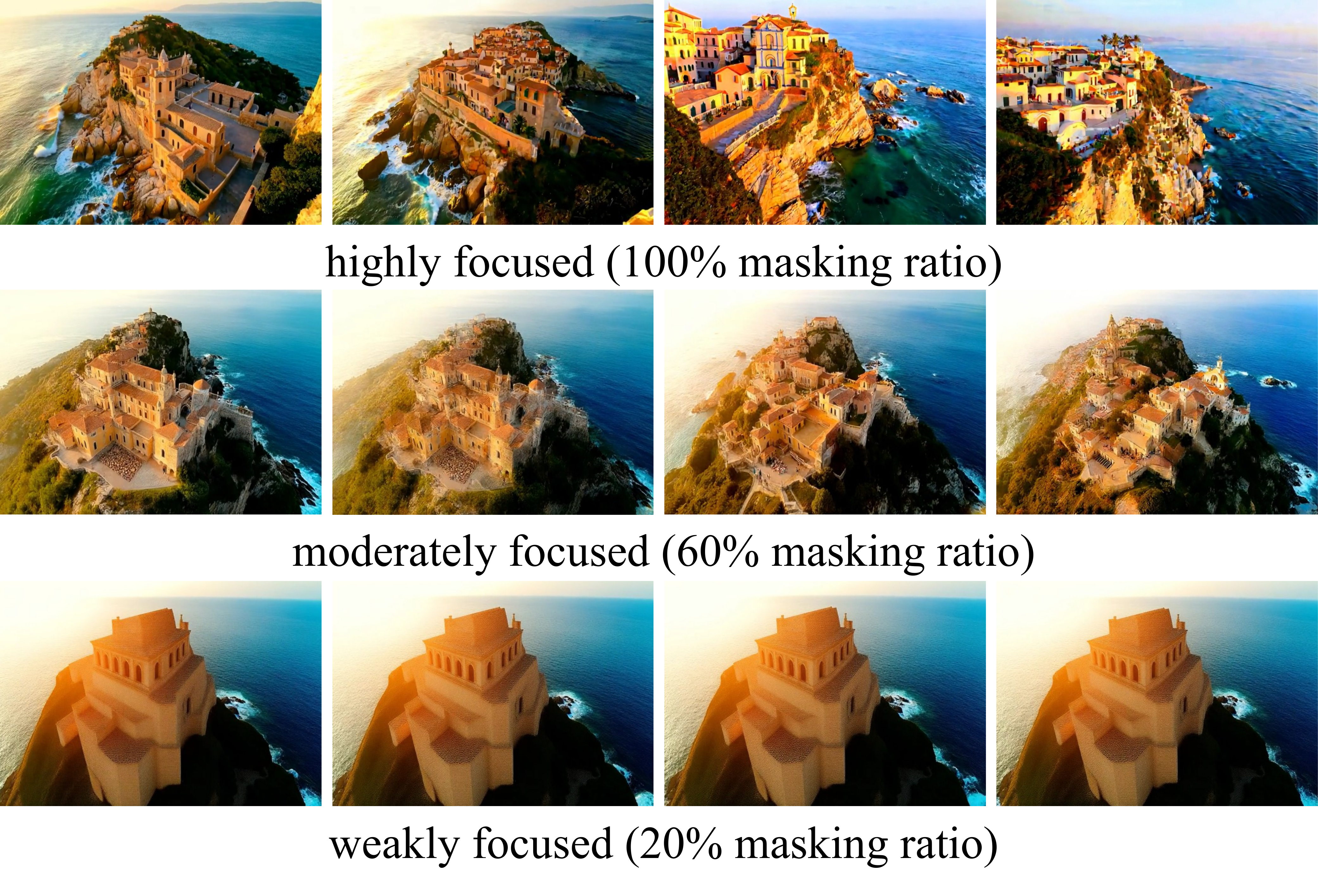}
        \subcaption{Qualitative results.}\label{fig:top-p-dropout:qual}
    \end{subfigure}
\vspace{-.25cm}
    \caption{
    % \textbf{Causal validation of attention dilution as the root cause of quality degradation.} (a) Quantitative and (b) qualitative results demonstrate a clear monotonic improvement in video quality and dynamics as the degree of attention focusing (i.e., the masking ratio of out-of-window scores) increases. 
    \textbf{Validation of attention dispersion as the cause of quality degradation.} Both (a) quantitative and (b) qualitative results show that video quality improves monotonically as the degree of attention central focusing (i.e., the masking ratio of out-of-window scores) increases.}
    \label{fig:top-p-dropout}
    \vspace{-0.5cm}
\end{figure}

 To validate this hypothesis, we conduct a controlled experiment where we progressively mask attention scores for tokens outside the training window, thereby forcing the attention to concentrate centrally. The results, presented in Fig.~\ref{fig:top-p-dropout}, demonstrate a clear positive correlation: more concentrated attention (i.e., by increasing the proportion of masked out-of-window scores) consistently improves both the visual quality and motion dynamics of the generated video. This provides strong evidence that attention dispersion is the cause of quality degradation. Consequently, as the extrapolation ratio increases, attention becomes more dispersed, leading to worse quality, consistent with the observations in Sec. ~\ref{sec: failure modes}.

% \begin{figure}
%     \centering
%     \includegraphics[width=\textwidth]{images/bottom-p-dropout/bottom-p-dropout.pdf}
%     \caption{\textbf{The relationship between the degree of attention focus within the training window and video quality.}  As the proportion of discarded attention outside the window increases, the attention focus within the window intensifies, leading to a simultaneous improvement in visual quality and dynamic performance of the videos.}
%     \label{fig:top-p droptout}
% \end{figure}

\textbf{A unified view: periodic attention as a case of attention dispersion.}~Building upon the above analysis, we can unify both failure modes under a single perspective: attention dispersion is the fundamental cause of extrapolation failure, with periodic attention patterns representing a special case. Specifically, when a RoPE frequency contributes substantially to the overall amplitude (e.g., due to harmonic alignment), it induces a strongly periodic attention pattern; otherwise, the model exhibits generic, non-periodic dispersion.

\subsection{UltraViCo}
\label{sec: method}

Building on the above unified view, we propose \emph{\textbf{Ultra}}-extrapolated \emph{\textbf{V}}ideo via Attention \emph{\textbf{Co}}ncentration (\emph{\textbf{UltraViCo}}), a simple yet effective method that suppresses attention for tokens beyond the training window via a decay factor, thereby restoring the model's focusing ability. To achieve this, we introduce a position-dependent decay factor $\lambda_{ij}$ applied to the original attention logits $S_{ij}$, yielding the corrected attention $S'_{ij}$:
\begin{equation}
S'_{ij} = \lambda_{ij} \cdot S_{ij}, \quad \text{where} \quad \lambda_{ij} =
\begin{cases}
1, & \text{if } |i-j| \le L/2 \text{ or } S_{ij} < 0, \\
\alpha, & \text{otherwise}, 
\end{cases}\\
\label{eq:focus_attention}
\end{equation}
where $\alpha<1$ is a constant decay hyperparameter and $L$ is the training length. Here, $\lambda_{ij}$ is set to be $1$ for all pairs within the training window, preserving the model's core learned dynamics. For out-of-window tokens, only positive logits ($S_{ij}\ge0$) are down-scaled because multiplying negative logits $S_{ij}<0$ by $\alpha<1$ can undesirably increase its value, while multiplying $\alpha > 1$ or 1 for negative logits has a negligible effect. We also experimented with various decay strategies, such as linear decay, but found the constant form is sufficient, indicating that the key is distinguishing in-window from out-of-window tokens rather than the decay shape itself (see Sec.~\ref{sec: experiments} for details).

However, in models showing periodic repetition (Sec.~\ref{sec:repetition_mechanism}), harmonic alignment positions $mT$ attract disproportionately high attention. Applying a uniform small decay $\alpha$ would overly suppress all out-of-window context, harming temporal consistency. To address this, we apply a stronger decay $\beta < \alpha$ specifically to these risky positions $mT$, while keeping $\alpha$ for other out-of-window tokens:
\begin{equation}
\label{eq:focus_attention_final}
\lambda_{ij} = 
\begin{cases} 
1, & \text{if } |i-j| \le L/2 \text{ or } S_{ij} < 0, \\
\beta, & \text{else if} \, (i,j) \in \mathcal{P}_{\text{risk}}, \\
\alpha, & \text{otherwise},
\end{cases}
\end{equation}
where $\mathcal{P}_{\text{risk}} = \{\, (i,j) | \  mT - \gamma \le i-j \le mT + \gamma,~ m\in \mathbb{Z},\gamma \in\mathbb{N}^+ \,\}$ denotes the set of positions within $\gamma$ frames around the harmonic alignment positions $mT$ and $\beta<\alpha<1$. This targeted adjustment reallocates attention to reliable in-window context while eliminating spurious periodic patterns, allowing UltraViCo to mitigate both failure modes simultaneously.

% 适当说一下细节：多长导致多大矩阵 oom。
 
\textbf{Efficient CUDA implementation.}
UltraViCo requires modifying attention logits, but standard PyTorch attention is infeasible for long sequences. At a $3\times$ extrapolation ($\sim$200K tokens for HunyuanVideo), for instance, materializing a $200\text{K} \times 200\text{K}$ attention mask consumes over $80$GB of memory in \texttt{bf16}, causing an immediate out-of-memory error. To address this, we integrate UltraViCo into Triton-based FlashAttention~\citep{dao2022flashattention} and SageAttention~\citep{zhang2024sageattention}, where the online-softmax formulation avoids explicit mask construction. 
This yields scalable, memory-efficient computation, enabling UltraViCo on large video models.

\section{Experiments}

\subsection{Setup}
\label{sec: setup}

\paragraph{Evaluation.} 
We evaluate methods on three video diffusion models, including HunyuanVideo, Wan2.1-1.3B and  CogVideoX-5B. Following RIFLEx, we use 100 prompts sampled from VBench~\citep{huang2024vbench}. For quantitative evaluation, following RIFLEx, we adopt Imaging Quality (Quality), Dynamic Degree (Dynamics), and Overall Consistency (Overall) from VBench, along with the NoRepeat Score for models prone to content repetition. Notably, our NoRepeat Score is a variant of that in RIFLEx, tailored for multiple-repetition  (see Appendix~\ref{appendix: more implement details} for details). Finally, we conduct a user study with 10 participants on 10 prompts, where users rank (User) the overall quality of videos across all methods. More details are provided in Appendix~\ref{appendix: more implement details}.

\textbf{Implementation Details.} 
The decay factor $\alpha$ is set to 0.9 for Wan and HunyuanVideo at $3\times$ and $4\times$ extrapolation. For HunyuanVideo, we set $\gamma=4$ for all ratios, and $\beta=0.6$ at $3\times$ and $0.8$ at $4\times$. Our baseline configurations follow RIFLEx. Further details are provided in Appendix~\ref{appendix: more implement details}.

\subsection{Results}
\label{sec: experiments}

\begin{table}[t!]
\centering
\caption{\textbf{Quantitative illustrative results on VBench for HunyuanVideo and Wan.} For Wan, which does not exhibit content repetition, we omit the NoRepeat Score. Additional results for more extrapolation ratios and models are provided in Appendix~\ref{appendix: more_quantitive_experiments}. Consist., Dyn., Qual., Over. and NoRe. denote Consistency, Dynamics, Quality, Overall and NoRepeat Score respectively. Normal. indicates the training length for reference.}
\label{tb: quantitive experiment}
\renewcommand\arraystretch{1}
\resizebox{\textwidth}{!}{
\begin{tabular}{lccccccccccc}
\toprule
\multirow{3}{*}{Method}  & \multicolumn{5}{c}{Wan2.1-1.3B} & \multicolumn{6}{c}{HunyuanVideo} \\
\cmidrule(r){2-6} \cmidrule(l){7-12} 
& \makecell{Consist.$\uparrow$} & \makecell{Dyn.$\uparrow$} & \makecell{Qual.$\uparrow$}  & \makecell{Over.$\uparrow$} & \makecell{User$\downarrow$} 
 & 
 \makecell{Consist.$\uparrow$} 
 & \makecell{NoRe.$\uparrow$} &\makecell{Dyn.$\uparrow$} & \makecell{Qual.$\uparrow$}  & \makecell{Over.$\uparrow$} & \makecell{User$\downarrow$} \\
 \midrule
Normal.  & 0.9554
& 51 & 70.34 & 24.25 & -- &  0.9786 & --  & 71  & 69.31 & 26.81 & -- 
 \\
\midrule
\multicolumn{12}{c}{$3\times$ extrapolation } \\
\midrule
PE  & 0.9419  & 6  & 56.28 & 18.53 & 3.82
& 0.9795 & 53.17 & 16 & 51.85 & 21.62 &  3.96
 \\
PI  & 0.9667  & 7  & 52.16 & 17.48 & 4.69
& 0.9787 & 90.23 & 1 & 46.30 & 21.29 &  4.91
 \\
NTK & 0.9437 & 3  & 57.73 & 18.50 &  4.40
& 0.9802 & 84.80 & 24 & 53.11 & 22.14 & 3.74
 \\
YaRN & \textbf{0.9676} & 5  & 53.46 & 17.53 &  4.71
& 0.9790 & 88.74 & 0 & 47.05 & 21.42 & 5.05
 \\
TASR & 0.9434 & 6  & 57.41 & 18.48 &  4.47
& 0.9807& 80.74 & 22 & 51.95 & 22.02 & 4.65
 \\
RIFLEx & 0.9431 & 5 & 53.79 & 17.54 & 4.90
& \textbf{0.9823} & 73.97 & 17 & 50.57 & 21.22 & 4.67
 \\
\textbf{Ours} & 0.944  &\textbf{46} & \textbf{62.43} & \textbf{23.21} & \textbf{ 1.01
}
& 0.9465& \textbf{100.0} & \textbf{62} & \textbf{65.00} & \textbf{26.45} &\textbf{ 1.02
} \\
\midrule
\multicolumn{12}{c}{$4\times$ extrapolation} \\
\midrule
PE  & 0.9415     & 11 & 55.25 & 16.65 &  3.75
& 0.9891 & 31.41 & 14 & 47.12 & 17.61 &  3.70
 \\
PI   & 0.9711   & 12 & 50.44 & 16.34 &  4.87
& 0.9885 & 70.93 & 0 & 42.19 & 17.83 &  4.82
 \\
NTK & 0.9477  & 11 & 55.37 & 16.09 & 4.24
& \textbf{0.9915} & 72.39 & 10 & 50.01 & 18.92 & 4.23
 \\
YaRN & \textbf{0.9729} & 7 & 51.16 & 16.69 & 4.57
& 0.9877 & 62.87 & 1 & 41.37 & 18.53 & 5.03
\\
TASR  & 0.9495 & 9 & 55.18 & 16.16 & 4.72
& 0.9911 & 51.28 & 14 & 46.81 & 18.47 & 4.51
 \\
RIFLEx & 0.9453  & 10 & 51.05 & 15.83 & 4.84
& 0.9906 & 52.84 & 11 & 41.02 & 16.47 & 4.69
 \\
\textbf{Ours} & 0.9484 & \textbf{47} & \textbf{59.36} & \textbf{21.61} & \textbf{ 1.01
}
& 0.9468& \textbf{99.87} & \textbf{42} & \textbf{66.54} & \textbf{24.52} & \textbf{ 1.02
}\\
\bottomrule
\vspace{-.3cm}
\end{tabular}
}
\vspace{-.3cm}
\end{table}

\begin{figure}[h!]
    \centering
    \includegraphics[width=\linewidth]{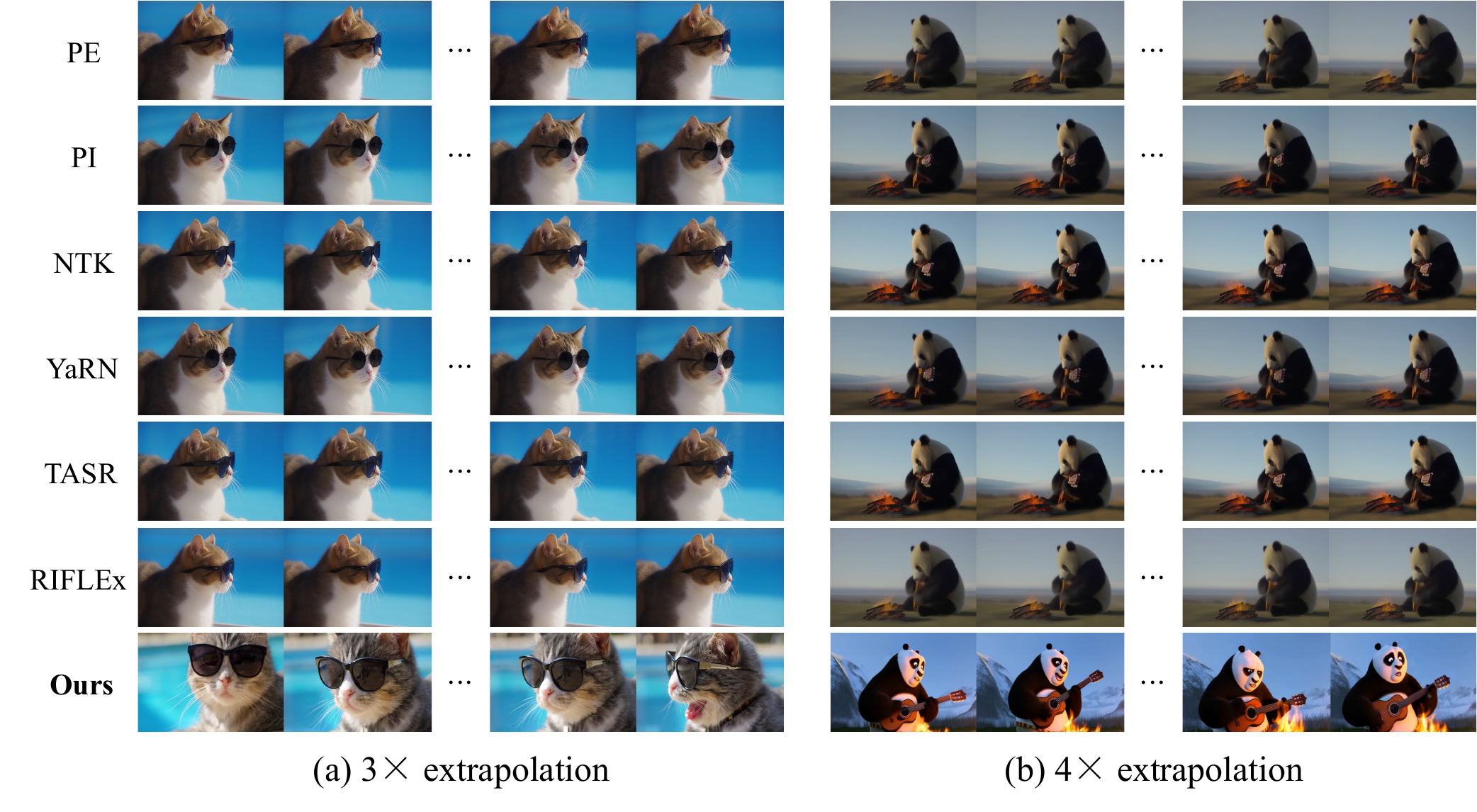}   
        \vspace{-.3cm}
        \caption{\textbf{Qualitative results on HunyuanVideo}. The baselines produce nearly static videos with poor visual quality, whereas our method achieves significantly better quality by addressing extrapolation failure modes. Additional qualitative results for other models are in Appendix~\ref{appendix: more_qualitive_experiments_wan_cog_3_4}.}
\vspace{-.3cm}
    \label{fig: 3x qualitative}
\end{figure}

\textbf{Performance comparison.} We compare a wide range of length extrapolation baselines on three SOTA models~\citep{kong2024hunyuanvideo, yang2024cogvideox, wan2025wan} across various extrapolation ratios, including PE~\citep{zhao2025riflex}, PI~\citep{chen2023extending}, NTK~\citep{bloc97}, TASR~\citep{zhuo2024lumina}, YaRN~\citep{peng2023yarn}, and RIFLEx. Tab.~\ref{tb: quantitive experiment} reports $3\times$ and $4\times$ results on HunyuanVideo and Wan, while Fig.~\ref{fig: 3x qualitative} shows qualitative samples on HunyuanVideo.  Results for additional ratios and models are provided in the Appendix~\ref{appendix: more_quantitive_experiments}.

As shown in Tab.~\ref{tb: quantitive experiment}, our method consistently outperforms all baselines across models and extrapolation ratios, simultaneously improving video quality and eliminating content repetition. Specifically, PE suffers from severe repetition,  reflected in low NoRepeat Scores.  In contrast, our method achieves substantially higher scores, effectively removing repetition. Beyond repetition, unlike RIFLEx which targets only this issue, our method delivers broader gains in both visual quality and motion quality. For instance, it improves Dynamic Degree and Imaging Quality on HunyuanVideo  by 233\% and 40.5\% over the previous best method at $4\times$ extrapolation, respectively. Notably, on Wan beyond $3\times$ extrapolation, while prior
methods collapse and yield static videos (Dynamic Degree $\le12$), our method restores fluid motion.
By addressing both core failure modes, our method extends the extrapolation limit from $2\times$ to $4\times$. These improvements are further corroborated by user rankings (Tab.~\ref{tb: quantitive experiment}) and qualitative visualizations (Fig.~\ref{fig: 3x qualitative}), which consistently confirm the superior quality of our generated videos over baselines.

\begin{figure}[h!]
  \centering
    \vspace{-.25cm}
  \begin{subfigure}
  {1.0\textwidth}
    \centering
    
      \centering
      \includegraphics[width=\linewidth,keepaspectratio]{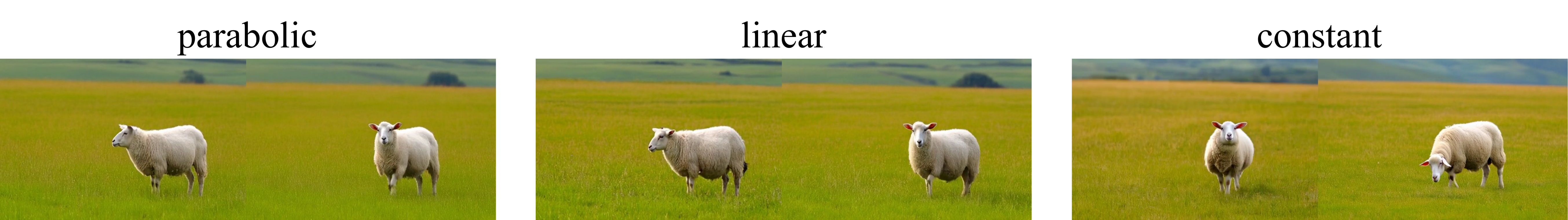}

    \label{fig:hunyuan}
  \end{subfigure}

  \begin{subfigure}
  {1.0\textwidth}
    \centering
    
      \centering
      \includegraphics[width=\linewidth,keepaspectratio]{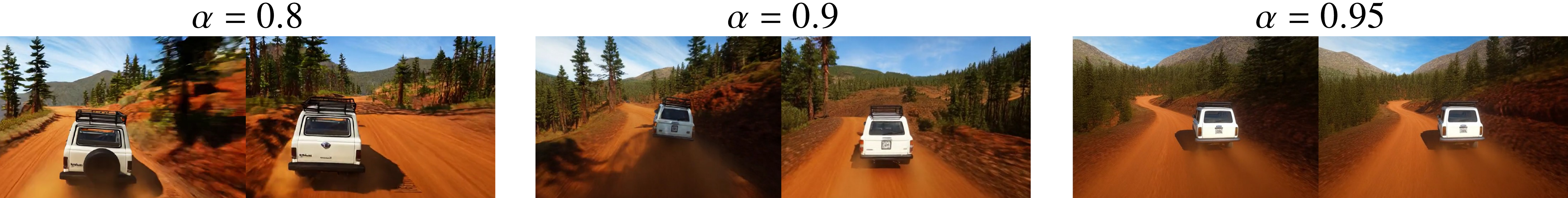}
      
   \label{fig:wan}
  \end{subfigure}

  \caption{\textbf{Ablation studies.} Top row: different decay strategies have minor impact, suggesting simple constant decay suffices. Bottom row: 
  small $\alpha$ harms consistency while large $\alpha$ offers limited gains. An intermediate value ($\alpha = 0.9$) enhances quality while preserving consistency.
}
  \label{fig:ablation}
% \vspace{-.5cm}
  
\end{figure}

\begin{figure}[h!]
  \centering
      \includegraphics[width=\linewidth,keepaspectratio]{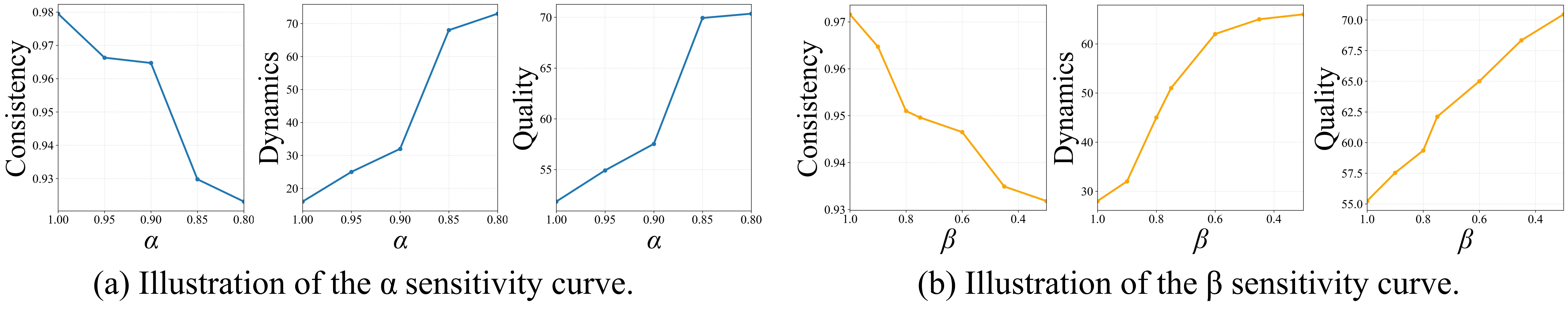}
  \vspace{-0.5cm}
  \caption{\textbf{Illustration of the hyperparameter sensitivity curve.} (a) When $\alpha \ge 0.9$, motion dynamics improve while consistency stays stable; below 0.9, consistency drops sharply. (b) When $\beta \ge 0.6$, dynamics remain high with comparable consistency; below $0.6$, consistency degrades significantly.}
  \label{fig:alphabet}

\end{figure}

\textbf{Ablation studies.} We ablate the decay strategy and the decay factor $\alpha$ on Wan at $3\times$ extrapolation. As shown in Fig.~\ref{fig:ablation} (top), different decay strategies yield minor differences, indicating that simple constant decay suffices. As shown in Fig.~\ref{fig:ablation} (bottom), strong decay harms consistency (i.e., the spare tire of the car disappears) while weak decay offers limited gains. An intermediate value ($\alpha=0.9$) enhances quality while preserving consistency. Further details are provided in Appendix~\ref{appendix: more implement details}. A sensitivity analysis for $\alpha$ and $\beta$ (Fig.~\ref{fig:alphabet}) shows a stable trend: $\alpha \ge 0.9$ and $\beta \ge 0.6$ improve visual quality and motion dynamics while keeping temporal consistency near baseline. We adopt $\alpha = 0.9$ and $\beta = 0.6$ as robust defaults, with small adjustments possible (e.g., $\beta=0.8$ for stronger consistency, $\alpha=0.85$ for better quality). Although larger $\alpha$ and $\beta$ may introduce a mild reduction in consistency, values above $0.94$ remain visually stable, aligning with common long-video settings (e.g., Wan’s training-horizon consistency $\approx 0.95$). See more metrics of $\alpha,\beta$ in Tab.~\ref{tb:rebuttal_alpha_sensitivity},~\ref{tb:rebuttal_beta_sensitivity},~\ref{tb:ablation_on_alphabet}, and Fig.~\ref{fig:appendix-alphabet}.

\textbf{Connection with other long-video generation methods.}
UltraViCo aims to extend the effective training window of video diffusion transformers and is therefore orthogonal to existing long-video generation techniques such as FreeNoise~\citep{qiu2023freenoise}, FIFO-Diffusion~\citep{kim2024fifo}, and sliding-window.
As demonstrated in Table~\ref{tb:other_long_video}, enlarging the context window via UltraViCo consistently improves the long-term temporal consistency of these methods, without negatively affecting other performance.
In Table~\ref{tb:other_long_video}, all methods follow the same evaluation setup ($6\times$ extrapolation for 30-second videos on Wan), where UltraViCo extends the base model’s training window by $3\times$.

\textbf{Generalization to downstream tasks.} Our method enhances the model’s inherent ability to handle longer sequences, making it naturally applicable to downstream tasks. As shown in Fig.~\ref{fig:ability}, based on VACE~\citep{jiang2025vace}, UltraViCo enables $3\times$ extrapolation in controllable generation and video editing. See Appendix~\ref{appendix: more_qualitive_experiments_wan_cog_3_4} for additional results.

\begin{table}[h!] \centering \caption{\textbf{Application of UltraViCo on existing long-video methods.}
} \label{tb:other_long_video} 
\renewcommand\arraystretch{1.0} \begin{tabular}{lcccc} \toprule \makecell{Method} & \makecell{Consistency$\uparrow$} & \makecell{Dynamics$\uparrow$} & \makecell{Quality$\uparrow$} & \makecell{Overall$\uparrow$} \\ \midrule Sliding Window & 0.8478 & 56 & 62.94 & 23.57\\ + UltraViCo & \textbf{0.9183} & 54 & 62.85 & 23.95 \\ \midrule FreeNoise & 0.9243 & 38 &63.09& 23.75\\ + UltraViCo & \textbf{0.9431} & 41 & 62.12 & 23.92\\ \midrule FIFO-Diffusion & 0.9131 & 53 & 61.31 & 23.81\\ + UltraViCo & \textbf{0.9319} & 51 & 63.09 & 24.24\\ \bottomrule \end{tabular} \end{table}

\section{Conclusion}

In this paper, we identify attention dispersion as the unified cause behind video length extrapolation failures. Based on this insight, we propose a training-free method that suppresses attention scores for tokens beyond training length. Experiments show that it significantly improves video quality, extending the practical extrapolation limit from $2\times$ to $4\times$. 

\section*{Ethics STATEMENT}
This paper advances the field of video generation, while emphasizing the importance of responsible use to avoid potential negative societal impacts, such as the creation of misleading or harmful content.

\section*{Acknowledgements}

This work was supported by Fundamental and Interdisciplinary Disciplines Breakthrough Plan of the Ministry of Education of China (No. JYB2025XDXM101); National Natural Science Foundation of China (Nos. 62522609, 92470118); the Beijing Natural Science Foundation (No. L247030); and the fund for building world-class universities (disciplines) of Renmin University of China.

% \section*{Reproducibility STATEMENT}
% Our code and the prompts in the paper are included in the supplementary material, and the implementation details are described in Sec.~\ref{sec: setup}. 

\bibliography{iclr2026_conference}
\bibliographystyle{iclr2026_conference}

\appendix

\section*{Use of Large Language Models}
We used a large language model solely to assist in polishing English writing and improving clarity. All research ideas, experiments, results, and interpretations are entirely our own.

\section{Related Work}
\label{sec: related work}
\paragraph{Text-to-video Diffusion Transformers.} Beyond autoregressive approaches~\cite{yan2025generation,Yan_2025_ICCV}, the recent advances in text-to-video generation have been primarily driven by diffusion models~\citep{ho2020denoising,song2020score,ho2022imagen,he2022lvdm,zhao2022egsde,zhao2023controlvideo,blattmann2023stable,xing2023dynamicrafter,chen2023videocrafter1,zhao2024identifying,polyak2024movie,zhou2024allegro,genmo2024mochi,chen2024videocrafter2,ma2024followpose,ma2024followyouremoji,ma2025controllable}. With the development of diffusion transformers (DiTs)~\citep{bao2023all,peebles2023scalable}, DiT-based text-to-video diffusion models have achieved remarkable performance, such as Sora \citep{videoworldsimulators2024}, Vidu \citep{bao2024vidu}, CogVideoX \citep{yang2024cogvideox} and Open-Sora \citep{zheng2024open}. Although achieving high quality, leading models are trained only on a fixed maximum sequence length, limiting long-term capacity. During video length extrapolation, they suffer from repetition or quality degradation, underscoring the need for length extrapolation.

\paragraph{Length Extrapolation in Transformers.} 

The goal of length extrapolation is to enable transformers to generate sequences longer than those seen during training in a single forward~\citep{press2021train}. This is typically achieved by modifying positional encodings.  For example, position interpolation (PI)~\citep{chen2023extending} improves performance by interpolating the frequencies in RoPE so that they remain within the training range even under extrapolation. NTK~\citep{bloc97}, YaRN~\citep{peng2023yarn}, and Time-aware Scaled RoPE (TASR)~\citep{zhuo2024lumina} combine interpolation with direct extrapolation, incorporating adjustments along the token dimension, denoising timesteps, and other factors to achieve better results. However, these methods perform poorly on image and video DiTs, often leading to content collapse or repetition. RIFLEx~\citep{zhao2025riflex} mitigates repetition by identifying and attenuating the intrinsic RoPE frequency, yet it still suffers from degraded visual quality. In contrast, our method effectively addresses both content repetition and quality degradation.

\paragraph{Long Video Generation.}
There also exist many approaches to long video generation~\citep{qiu2023freenoise,wang2023gen,henschel2025streamingt2v,kim2024fifo,tan2024video,yin2025slow,wang2024loong,cai2025ditctrl,li2025longdiff,lu2024freelong,tan2025freepca,jiang2025lovic,gao2025longvie,gu2025long}, most of which intervene in the diffusion inference process. For instance, FreeNoise~\citep{qiu2023freenoise} enhances temporal consistency via noise initialization, FIFO-Diffusion~\citep{kim2024fifo} feeds frames sequentially into a denoising window of training length, and Video-Infinity~\citep{tan2024video} exploits distributed computation to scale up video length. While effective for generating long videos, these methods are orthogonal to our length extrapolation strategy, which extends the intrinsic capacity of DiTs to longer sequences and can be readily integrated with them.

In addition to diffusion-based approaches to long video generation, alternative modeling paradigms such as autoregressive methods~\citep{wu2021godiva,yan2021videogpt,hong2022cogvideo,wu2022nuwa,kondratyuk2023videopoet,wu2024vila,sun2024generative,wang2024emu3} and diffusion forcing~\citep{chen2024diffusion,huang2025self,teng2025magi} are also capable of generating long videos. Although our method is designed for diffusion models, it may also offer insights into length extrapolation for these alternative paradigms.

\section{More Details of Our Method}
\label{appendix: more details of the method}
\subsection{Derivation of the Periodic Outputs}
\label{appendix: derivation of the periodic outputs}
In this section, we present a formal derivation of Eq.~(\ref{eq: periodic attention}). Specifically, the attention score matrix $\mP\in\mathbb{R}^{L^\prime\times L^\prime}$ satisfies the following properties up to negligible error:

\textbf{Prop.1 }(Row-wise periodicity): $\mP_{i,j}=\mP_{i,j+T}, \forall i\in \{0,\dots,L^\prime-1\}, j \in \{0,\dots, L^\prime-T-1\},$ where $T\in\mathbb{N}^+$ corresponds to the observed repetition period in Sec.~\ref{sec: failure modes}. 

\textbf{Prop.2} (Relative positional invariance): $\mP_{i,j}=\mP_{i+p,j+p}, \forall i\in \{0,\dots,L^\prime-p-1\}, j \in \{0,\dots, L^\prime-p-1\},$ where $p\in\mathbb{N}^+$ is the relative displacement. In the ffollowing derivation we instantiate $p=T$.

On basis of the above properties, we derive the periodicity of the attention scores and outputs as follows. $\forall i \in \{0,\dots,L^\prime-T-1\},$
\begin{align}
    \mO_{i+T} &= \quad\sum\nolimits_{j=0}^{L^\prime-1}\mP_{i+T,j}\mV_j\\
    &=\quad
    \sum\nolimits_{j=0}^{L^\prime-T-1}\mP_{i+T,j}\mV_j+\sum\nolimits_{j=L^\prime-T}^{L^\prime-1}\mP_{i+T,j}\mV_j\\
    &\eqby{\text{Prop.1 }}\quad
    \sum\nolimits_{j=0}^{L^\prime-T-1}\mP_{i+T,j+T}\mV_j+\sum\nolimits_{j=L^\prime-T}^{L^\prime-1}\mP_{i+T,j}\mV_j\\
    &\eqby{\text{Prop.2 }}\quad
    \sum\nolimits_{j=0}^{L^\prime-T-1}\mP_{i,j}\mV_j+\sum\nolimits_{j=L^\prime-T}^{L^\prime-1}\mP_{i,j-T}\mV_j\\
    &\eqby{\text{Prop.1 }}\quad
    \sum\nolimits_{j=0}^{L^\prime-T-1}\mP_{i,j}\mV_j+\sum\nolimits_{j=L^\prime-T}^{L^\prime-1}\mP_{i,j}\mV_j\\
    &=\quad\sum\nolimits_{j=0}^{L^\prime-1}\mP_{i,j}\mV_j\\
    &=\quad\mO_i.
\end{align}
\subsection{Details of the Multimodal Rotary Position Embedding}
\label{appendix: M-RoPE}
In this section, we provide the details of the Multimodal RoPE (M-RoPE)~\citep{wang2024qwen2} introduced in Sec.~\ref{sec:rope}. Specifically, for a token at position $(t,h,w)$, the input vector $\vx \in \R^D$ is divided into three subspaces of dimensions $d_\gT, d_\gH, d_\gW$, respectively assigned to temporal, height, and width encodings. Each subspace is modulated by its own frequency series $\{\phi_i^\gT\}_{i=0}^{d_\gT-1}, \{\phi_i^\gH\}_{i=d_\gT}^{d_\gT+d_\gH-1}, \{\phi_i^\gW\}_{i=d_\gT+d_\gH}^{D-1}$. Concretely, we define
\begin{align}
\boldsymbol{f}^{\text{RoPE}}(\vx,t,h,w)_i
= R_i^{\alpha}(p_\alpha)
\begin{bmatrix}
   x_{2i} \\
   x_{2i+1}
\end{bmatrix}, 
\quad
R_i^{\alpha}(p_\alpha) =
\begin{bmatrix}
   \cos (\phi_i^\alpha p_\alpha) & -\sin  (\phi_i^\alpha p_\alpha) \\
   \sin (\phi_i^\alpha p_\alpha ) &  \cos (\phi_i^\alpha p_\alpha )
\end{bmatrix},
\end{align}
where $\alpha \in \{\gT,\gH,\gW\}$ indexes the temporal, height, and width dimensions with corresponding positions $p_\alpha \in \{t,h,w\}$ and frequency components $\{\phi_i^\alpha\}$. The index ranges are
\begin{align}
i \in 
\begin{cases}
\{0,\dots,d_\gT/2-1\}, & \alpha = \gT,\\
\{d_t/2,\dots,d_\gT/2+d_\gH/2-1\}, & \alpha = \gH,\\
\{d_\gT/2+d_\gH/2,\dots,D/2-1\}, & \alpha = \gW.
\end{cases}
\end{align}
After M-RoPE encoding, the queries and keys form $\mQ \in \mathbb{R}^{L' \times D}$ and $\mK \in \mathbb{R}^{L' \times D}$. As in Eq.~(\ref{eq:attention}), they produce the attention logits matrix $\mS \in \mathbb{R}^{L' \times L'}$, where the attention logit between the query at $(t,h,w)$, denoted $q_{(t,h,w)}$, and the key at $(t+\Delta t,h+\Delta h,w + \Delta w)$, denoted $k_{(t+\Delta t,h+\Delta h,w + \Delta w)}$, expands explicitly as:
\begin{align}
\label{eq: app-m-rope}
\mS_{(t,h,w),(t+\Delta t,h+\Delta h,w + \Delta w)} &=\sum_{i=0}^{d_\gT/2-1} q_{(t,h,w)}^{(2i: 2i+1)\top} \mR_{i}^{\gT}(\Delta t)k_{(t+\Delta t,h+\Delta h,w + \Delta w)}^{(2i: 2i+1)}+  \notag
\\&
\sum_{i=d_\gT/2}^{d_\gT/2+d_\gH/2-1} q_{(t,h,w)}^{(2i: 2i+1)\top} \mR_i^{\gH}(\Delta h)k_{(t+\Delta t,h+\Delta h,w + \Delta w)}^{(2i: 2i+1)}+ \notag\\
&\sum_{i=d_\gT/2+d_\gH/2}^{D/2-1} q_{(t,h,w)}^{(2i: 2i+1)\top} \mR_i^{\gW}(\Delta w)k_{(t+\Delta t, h+\Delta h, w+\Delta w)}^{(2i: 2i+1)}\\
&=\sum_{i=0}^{d_\gT/2-1}\Big[\lambda_1^{(i)}\cos(\phi_i^\gT\Delta t)+\lambda_2^{(i)}\sin(\phi_i^\gT\Delta t)\Big]+  \notag
\\& \sum_{i=d_\gT/2}^{d_\gT/2+d_\gH/2-1}\Big[\lambda_1^{(i)}\cos(\phi_i^\gH\Delta h)+\lambda_2^{(i)}\sin(\phi_i^\gH\Delta h)\Big]+ \notag\\
&
\sum_{i=d_\gT/2+d_\gH/2}^{D/2-1}\Big[\lambda_1^{(i)}\cos(\phi_i^\gW\Delta w)+\lambda_2^{(i)}\sin(\phi_i^\gW\Delta w)\Big],
\end{align}
where 
\begin{align}
    \lambda_1^{(i)} = q_{(t, h, w)}^{(2i)}k_{(t+\Delta t, h+\Delta h, w+\Delta w)}^{(2i)}+q_{(t, h, w)}^{(2i+1)}k_{(t+\Delta t, h+\Delta h, w+\Delta w)}^{(2i+1)},\\
    \lambda_2^{(i)} = q_{(t, h, w)}^{(2i+1)}k_{(t+\Delta t, h+\Delta h, w+\Delta w)}^{(2i)}-q_{(t, h, w)}^{(2i)}k_{(t+\Delta t, h+\Delta h, w+ \Delta w)}^{(2i+1)}.
\end{align}
\subsection{Derivation of the Statistical Attention Pattern $\bar{\mS}(\Delta t)$ }
\label{appendix: derivation of expectation}
In this section, we present the derivation of Eq.~(\ref{eq:trig_decomposition}) in Sec.~\ref{sec:repetition_mechanism}. We investigate the row-wise pattern of attention logits by examining the expectation of the attention logits between queries and keys at relative temporal distance $\Delta t$ (i.e., $\E\big[\mS_{(t, h, w), (t+\Delta t, h, w)}\big])$\footnote{Strictly speaking, the analysis should target $\mS_{(t, h, w), (t+\Delta t, h+\Delta h, w+\Delta w)}$ for all $\Delta h, \Delta w$, but as the phenomena are similar across $\Delta h, \Delta w$, we focus on $\mS_{(t, h, w), (t+\Delta t, h, w)}$ for simplicity.}.
This expectation is taken across attention layers, heads, and query positions. In Appendix~\ref{appendix: actual qk variance}, we further show that when the true variance is taken into account, the actual attention logits still follow the same patterns as indicated by this expectation.

Specifically, on basis of the formula of M-RoPE (i.e., Eq. (\ref{eq: app-m-rope})), the target expectation is given by\footnote{For brevity, we omit layer and head indices in the expectation notation.}
\begin{align}
   & \E_{t, h, w}\Big[\mS_{(t, h, w), (t+\Delta t, h, w)}\Big] = \E_{t, h, w}\Big[\sum_{i=0}^{d_\gT/2-1} q_{(t, h, w)}^{(2i: 2i+1)\top} \mR_{i}^{\gT}(\Delta t)k_{(t+\Delta t, h, w)}^{(2i: 2i+1)} +\notag\\
&\sum_{i=d_\gT/2}^{d_\gT/2+d_\gH/2-1} q_{(t, h, w)}^{(2i: 2i+1)\top} \mR_i^{\gH}(0)k_{(t+\Delta t, h, w)}^{(2i: 2i+1)}+\sum_{i=d_\gT/2+d_\gH/2}^{D/2-1} q_{(t, h, w)}^{(2i: 2i+1)\top} \mR_i^{\gW}(0)k_{(t+\Delta t, h, w)}^{(2i: 2i+1)}\Big]\\
&= \sum_{i=0}^{d_\gT/2-1}\Big[
E_1^{(i)}\cos\big(\phi_{i}^{\gT}\Delta t\big) +E_2^{(i)}\sin\big(\phi_{i}^{\gT}\Delta t\big) \Big] + \sum_{i=d_\gT/2}^{D/2-1}E_1^{(i)},
\end{align}
where
\begin{align}
\label{eq:22}
    E_1^{(i)} = \E_{ t,h,w}\Big[q_{(t, h, w)}^{(2i)}k_{(t+\Delta t, h, w)}^{(2i)}+q_{(t, h, w)}^{(2i+1)}k_{(t+\Delta t, h, w)}^{(2i+1)}\Big],\\
\label{eq:23}
    E_2^{(i)} = \E_{ t,h,w}\Big[q_{(t, h, w)}^{(2i+1)}k_{(t+\Delta t, h, w)}^{(2i)}-q_{(t, h, w)}^{(2i)}k_{(t+\Delta t, h, w)}^{(2i+1)}\Big].
\end{align}
In practice, though the integrands of these expectations are actually functions of $\Delta t$, the empirical statistics in Fig.~\ref{fig: supplement_hy_wan} (col.~1) indicate that their variances with respect to $\Delta t$ are negligible. Hence, we approximate $E^{(i)}_1$ and $E^{(i)}_2$ as constants up to negligible error, which is defined by
\begin{align}
\label{eq: hat_e_1}
E_1^{(i)} \approx \mathbb{E}_{ t, h, w, \Delta t}
\Big[q_{(t, h, w)}^{(2i)}k_{(t+\Delta t, h, w)}^{(2i)}
+q_{(t, h, w)}^{(2i+1)}k_{(t+\Delta t, h, w)}^{(2i+1)}\Big]=:\hat{E}_1^{(i)} ,\\
\label{eq: hat_e_2}
E_2^{(i)} \approx \mathbb{E}_{ t, h, w, \Delta t}
\Big[q_{(t, h, w)}^{(2i+1)}k_{(t+\Delta t, h, w)}^{(2i)}
-q_{(t, h, w)}^{(2i)}k_{(t+\Delta t, h, w)}^{(2i+1)}\Big]=:\hat{E}_2^{(i)} .
\end{align}
By substituting these two expressions into Eq.~(\ref{eq:22}) and Eq.~(\ref{eq:23}), the expected attention logits can be well approximated as $\bm{\bar{S}}(\Delta t)$, where
\begin{align}
    \bm{\bar{S}}(\Delta t)
     =\sum_{i=0}^{d_\gT/2-1}\Big[
\hat{E}_1^{(i)}\cos\big(\phi_i^{\gT}\Delta t\big) +\hat{E}_2^{(i)}\sin\big(\phi_i^{\gT}\Delta t\big) \Big] +\sum_{i=d_\gT/2}^{D/2-1}\hat{E}_1^{(i)}.
\end{align}
To simplify the expression, we employ the auxiliary angle formula to rewrite the two trigonometric functions as one, i.e.,
\begin{align}
\bm{\bar{S}}(\Delta t)
     = \sum_{i=0}^{d_\gT/2-1}\Big[a_i \cos(\phi_i\Delta t+b_i)\Big] +C,
\end{align}
where $a_i = \sqrt{\Big[\hat{E}_1^{(i)}\Big]^2+\Big[\hat{E}_2^{(i)}\Big]^2},b_i = \text{atan2}(-\hat{E}_2^{(i)},\hat{E}_1^{(i)})$.
Interestingly, as shown in Fig.~\ref{fig: supplement_hy_wan} (col.~2), $\hat{E}_2^{(i)}$ remains consistently close to zero, which in turn makes $b_i$ nearly vanish (for example, $b_0$ is $0.039$ for HunyuanVideo). This observation allows us to apply Proposition~\ref{prop:periodicity} in Sec.~\ref{sec:repetition_mechanism} up to an error of negligible magnitude. Detailed statistical data for $\hat{E}_1^{(i)}, \hat{E}_2^{(i)}, a_i, b_i$ are shown in Fig.~\ref{fig: supplement_hy_wan} (col.~2, 3, 4).

\begin{figure}[h!]
  \centering

  \begin{subfigure}[h!]{\textwidth}
    \centering
    
      \centering
      \includegraphics[width=\linewidth,keepaspectratio]{appendix_imgs/hy.pdf}
      \subcaption{Statistics of HunyuanVideo.}
  \end{subfigure}

  \begin{subfigure}[h!]{\textwidth}
    \centering
    
      \centering
      \includegraphics[width=\linewidth,keepaspectratio]{appendix_imgs/wn.pdf}
      \subcaption{Statistics of Wan.}
  \end{subfigure}

  \caption{\textbf{Statistics of attention logits in HunyuanVideo and Wan.} The variances of $E_1^{(i)}, E_2^{(i)}$ with respect to $\Delta t$ (col.~1) are negligible compared to their expectations (col.~2), making the approximation in Eq.~(\ref{eq: hat_e_1}), Eq.~(\ref{eq: hat_e_2}) accurate. The bias angles $b_i$ (col.~4) are close to zero, except for $b_9$ and $b_{15}$ in Wan whose impact is negligible since the corresponding $a_9, a_{15}$ are near zero (col.~3).
}
  \label{fig: supplement_hy_wan}

\end{figure}

\subsection{Consistency of Actual Attention Pattern with $\bar{\mS}(\Delta t)$}
\label{appendix: actual qk variance}
\begin{figure}[h!]
    \centering
    \includegraphics[width=0.75\linewidth]{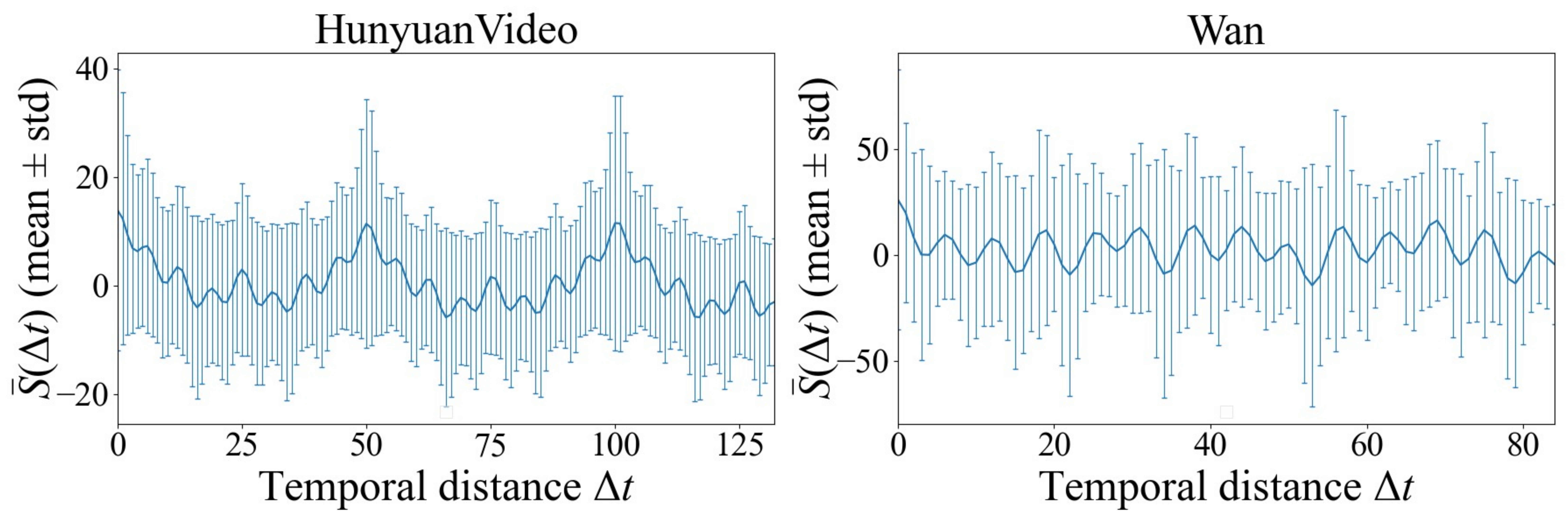}   \caption{\textbf{Attention logits under actual variance.} Even with standard deviation across layers, heads, and query positions, HunyuanVideo retains clear periodic peaks while Wan 2.1 remains non-periodic, confirming the general validity of the mean-based analysis in Sec.~\ref{sec:repetition_mechanism}.}
    \label{fig: errbar}
\end{figure}

In this section, we investigate the actual attention scores under the true variance, demonstrating that they preserve the same characteristics as the averaged values described in Sec.~\ref{sec:repetition_mechanism}. As shown in Fig.~\ref{fig: errbar}, when the standard deviation over attention layers, heads, and query positions is incorporated into the mean, the attention logits of HunyuanVideo still exhibit clear periodicity at their peaks, whereas those of Wan2.1 remain non-periodic. Therefore, the conclusions drawn in Sec.~\ref{sec:repetition_mechanism} from the mean-based analysis hold with strong generality in practice.

\subsection{Proof of Proposition~\ref{prop:periodicity}} 
Proposition~\ref{prop:periodicity} is well-known in harmonic analysis and signal processing, and we provide the proof here only for completeness.
\begin{proof}

\textbf{Sufficiency.} If $\phi_i/\phi_{N-1}\in\mathbb{N}^+$ for all $i$, write $\phi_i=k_i\phi_{N-1}$ with $k_i\in\mathbb{N}^+$. Let $T_{N-1}=2\pi/\phi_{N-1}$. Then for each $i$,
\begin{align}
\cos\big(\phi_i(\Delta t+T_{N-1})\big)
=\cos\!\big(k_i\phi_{N-1}\Delta t+2\pi k_i\big)
=\cos(\phi_i\Delta t),\quad \forall \Delta t\in\R,
\end{align}
so $f(\Delta t+T_{N-1})=f(\Delta t),\,\forall \Delta t\in\R$. Hence $T_{N-1}$ is a period of $f$.

\textbf{Necessity.}
Suppose $T_{N-1}=2\pi/\phi_{N-1}$ is a period of $f$. Then for all $\Delta t$, 
\begin{align}
    0=f(\Delta t+T_{N-1})-f(\Delta t)
=\sum_{i=0}^{N-1} a_i\big[\cos(\phi_i\Delta t+\phi_i T_{N-1})-\cos(\phi_i\Delta t)\big].
\end{align}
Using $\cos(x+y)-\cos x=(\cos y-1)\cos x-\sin y\,\sin x$,
\begin{align}
    0=\sum_{i=0}^{N-1} a_i\Big[(\cos(\phi_i T_{N-1})-1)\cos(\phi_i\Delta t)-\sin(\phi_i T_{N-1})\sin(\phi_i\Delta t)\Big],\quad\forall\,\Delta t\in\R.
\end{align}
The family $\{\cos(\phi_i\cdot),\sin(\phi_i\cdot)\}_i$ with distinct positive $\phi_i$ is linearly independent over $\mathbb{R}$ (e.g., via independence of $e^{\pm i\phi_i t}$). Hence for each $i$,
\begin{align}
    \cos(\phi_i T_{N-1})-1=0,\qquad \sin(\phi_i T_{N-1})=0,
\end{align}
so $\phi_i T_{N-1}\in 2\pi\mathbb{Z}$. Substituting $T_{N-1}=2\pi/\phi_{N-1}$ yields
\begin{align}
    \frac{\phi_i}{\phi_{N-1}}\in\mathbb{N}^+,
\end{align}
as all $\phi_i>0$. 
\end{proof}

\subsection{Remarks on Proposition~\ref{prop:periodicity}}
\label{appendix: lemma remarks}

\begin{figure}
    \centering
    \resizebox{\textwidth}{!}{
    \begin{tabular}{c|c|c}
    \toprule
    \scriptsize \small \textbf{Model} &\small \makecell{\textbf{Attention maps} } & 
    \small \makecell{\textbf{Statistical row attention analysis}}\\ \midrule 
\multirow{2}{*}{\makecell[t]{\small \textbf{Hun.}}} &
    
    \begin{minipage}{0.2\textwidth}
    \centering
\includegraphics[width=\textwidth]{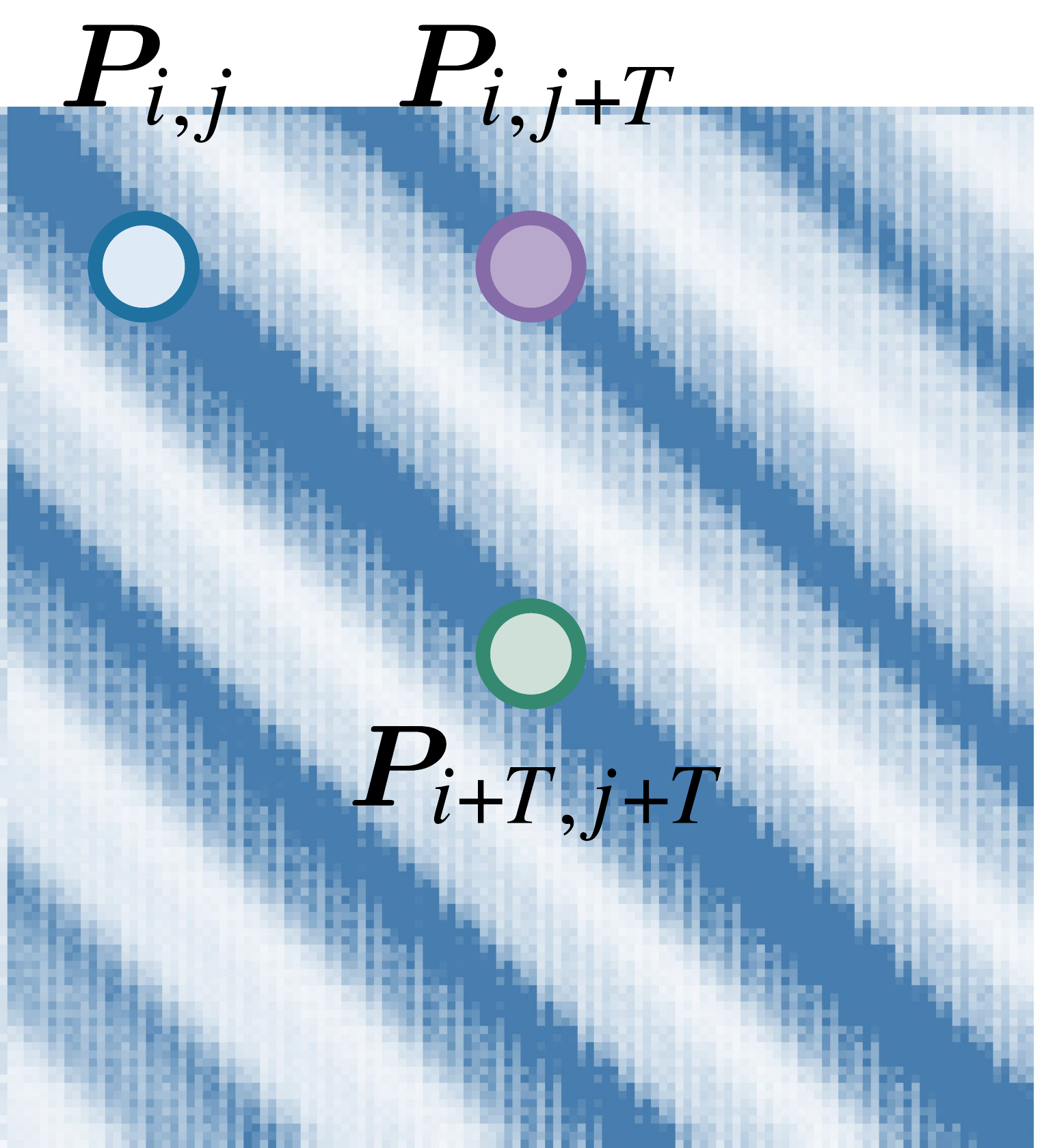}

    \end{minipage}&
    \begin{minipage}{0.8\textwidth}
    \centering   
    \includegraphics[width=\textwidth]{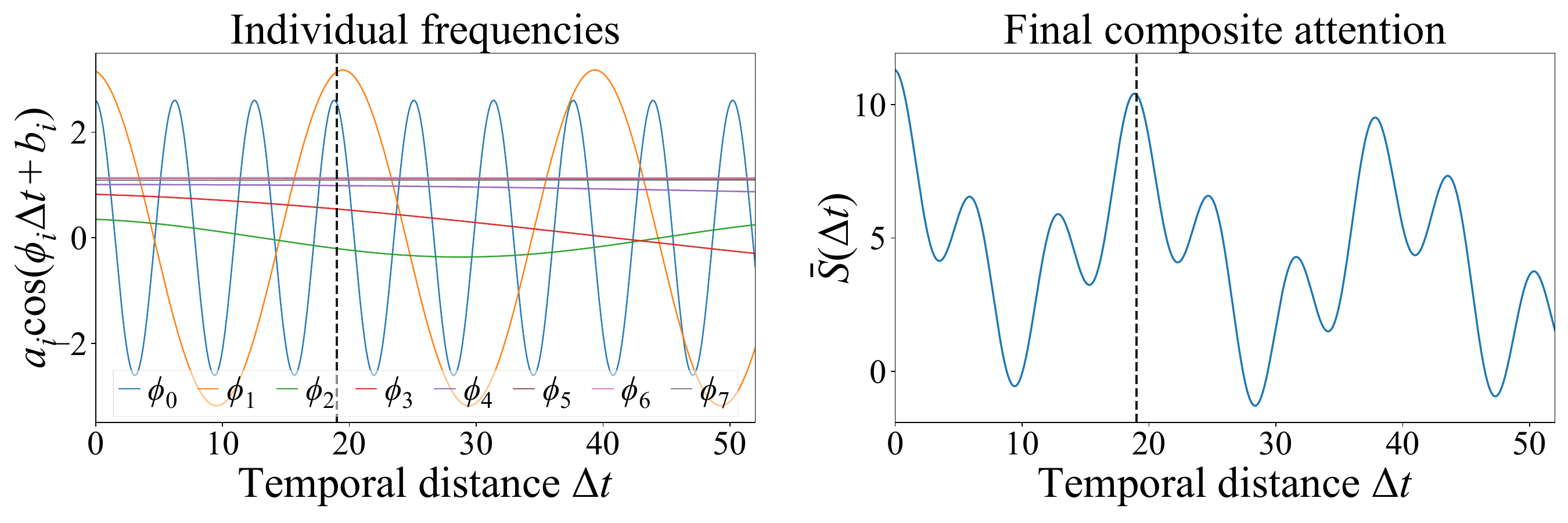}
    
    \end{minipage}
    \\
    &\small{(a) Periodic attention:}
    & \small{(b) Approximately harmonic RoPE frequencies ($\phi_0/\phi_1 \approx \mathbb{N}^+$) amplify the largest
    }\\
    &\small{$\mP_{i,j}\approx\mP_{i,j+T}$}
    & \small{amplitude $\phi_1$ (dashed line), inducing approximately periodic composite attention.}\\
     \bottomrule
    \end{tabular}
    }
    \caption{\textbf{Periodic attention patterns of CogVideoX.} The RoPE frequencies of CogVideoX approximately satisfy the harmonic condition, which amplifies the largest-amplitude component and thereby induces periodic attention patterns.}
    \label{fig: cogvideo repeat}

\end{figure}

\paragraph{Relaxed conditions under which the proposition holds approximately.} 
Although the strict condition for forming harmonics in Proposition~\ref{prop:periodicity} is $\phi_i/\phi_{N-1}\in\mathbb{N}^+$, in this section we highlight approximate conditions that can likewise induce a dominant frequency leading to content repetition in videos. Specifically, if $\phi_i/\phi_{N-1}$ is sufficiently close to an integer, constructive amplification can still occur for small $|t|$ (e.g., $|t|\le 2T_{N-1}$). For example, for CogVideoX, the ratio of the first two frequencies is $\phi_0/\phi_1=3.16$, which is close to the integer 3, thereby producing a dominant component that accounts for 50.80\% of the total amplitude. This gives rise to an approximately periodic composite attention pattern (Fig.~\ref{fig: cogvideo repeat}), which in turn leads to content repetition (Fig.~\ref{fig: cog-failure}, right).
\renewcommand\cellset{\renewcommand\arraystretch{0.7}}
\paragraph{Remarks on the strict period of HunyuanVideo.} 
We herein examine the strict periodicity of HunyuanVideo. Strictly speaking, its fundamental frequency is $\phi_7$, with ratios $\phi_i/\phi_7 = 2^{7-i}, i\in\{0,\dots,7\}$. According to Proposition \ref{prop:periodicity}, the theoretical period of $\bm{\bar{\mS}}(\Delta t)$ is $T_7 = \tfrac{2\pi}{\phi_7}$. However, as shown in Fig.~\ref{fig: supplement_hy_wan}a (col.~3), the amplification contributed by $\phi_7$ is very small, accounting for only 6.677\%, which makes its impact negligible. Moreover, its period of 804 is far larger than the extrapolation length (e.g., 132 at $4\times$ extrapolation), rendering the variation of the corresponding component almost imperceptible within this range. The same reasoning applies to $\phi_i$ for $i \in \{4,5,6\}$. Consequently, our analysis focuses on $\phi_i$ with $i \in \{0,1,2,3\}$, whose single-frequency contributions are both large enough in amplitude and sufficiently oscillatory to shape $\bm{\bar{\mS}}(\Delta t)$.

\subsection{Necessity of Concentrating on the Training Window}
\label{appendix: training window}
\begin{figure}[h!]
    \centering
    \includegraphics[width=1\linewidth]{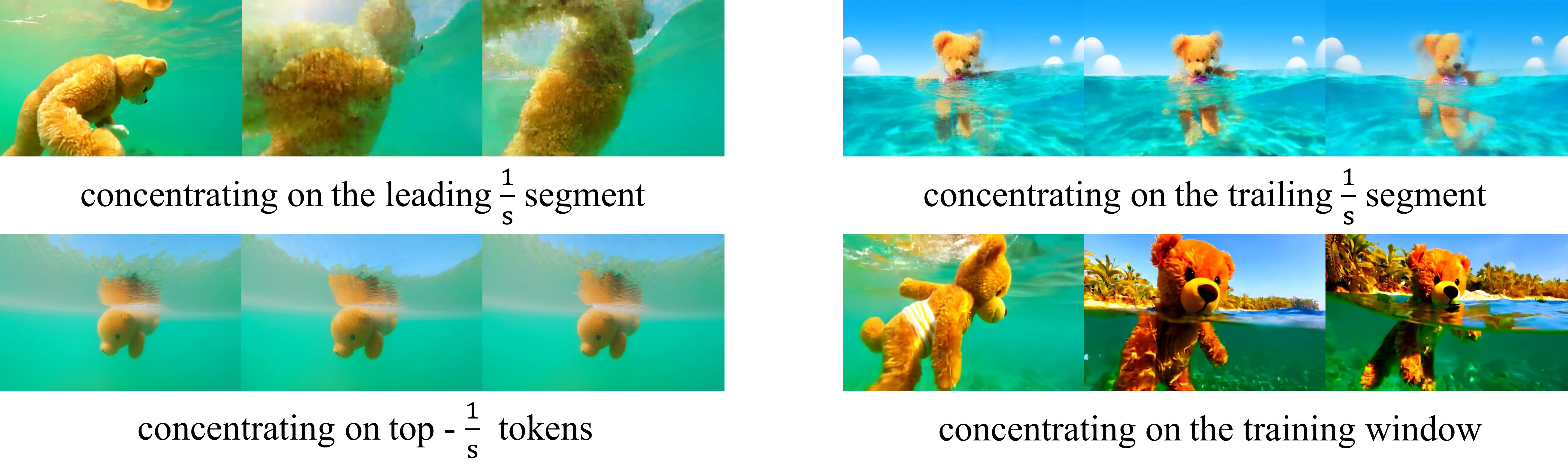}   \caption{\textbf{Comparison of attention concentration strategies on Wan at $s=3$.} Concentrating on the leading or trailing $\frac{1}{s}$ of each row collapses the video, and top--$\frac{1}{s}$ yields poor quality with little dynamics. Restricting attention to the training window proves most effective.}
    \label{appfig: training window}
\end{figure}

In this section, we provide detailed experimental evidence supporting the discussion in Sec.~\ref{sec: dispersion} on where sharpened attention focus is most beneficial. Specifically, on Wan with extrapolation ratio $s=3$, we test four strategies for sharpening attention: concentrating on the leading $\frac{1}{s}$ of each row, the trailing $\frac{1}{s}$, the training window, and the top--$\frac{1}{s}$ tokens according to the original attention scores. As shown in Fig. \ref{appfig: training window}, concentrating on the leading or trailing $\frac{1}{s}$ of each row causes the video to collapse, while top--$\frac{1}{s}$ yields poor visual quality with little dynamics. In contrast, restricting attention to the training window leads to the most significant improvement in video quality.

\section{More Details of Experiments}
\subsection{Failure Modes of CogVideoX}
\label{appendix:faliure-modes-of-CogVideoX}
\begin{figure}[h!]
    \centering
    \includegraphics[width=1\linewidth]{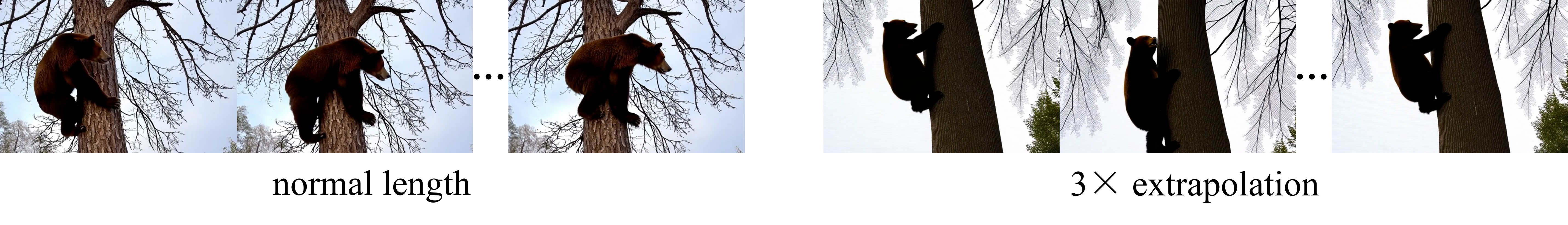}   \caption{\textbf{Failure modes of CogVideoX under $3\times$ extrapolation. }The generated videos show degraded visual quality, reduced dynamics, and clear content repetition, consistent with the failure modes discussed in Sec.~\ref{sec: failure modes}.}
    \label{fig: cog-failure}
\end{figure}

In this section, we present the manifestation of the failure modes of video length extrapolation as discussed in Sec.~\ref{sec: failure modes} on an additional model, CogVideoX. As shown in Fig.~\ref{fig: cog-failure}, when extrapolated to three times the normal training length, the generated videos exhibit a sharp decline in both dynamic degree and visual quality, along with noticeable content repetition.

\subsection{More Implementation Details}
\label{appendix: more implement details}
In this section, we provide further details of Sec.~\ref{sec: experiments}.
\paragraph{The implementation of NoRepeat Score.}
The NoRepeat Score implemented in RIFLEx~\citep{zhao2025riflex} is only applicable when the content repeats once, which makes it unsuitable for longer extrapolation tasks. We therefore modify it accordingly. Specifically, the computation of the NoRepeat Score consists of two steps: static-video filtering and repeated-frame ratio calculation. In the first step, we uniformly sample 8 frames across the video; if the mean pairwise $L_2$ distance among them falls below a threshold, the video is considered static and discarded. This prevents completely static videos from interfering with subsequent repetition detection. In the second step, we measure the ratio of repeated frames to the total frame count, which defines the NoRepeat Score. Following RIFLEx, we first search around the dominant-frequency period for the frame with the minimal $L_2$ distance to the first frame. This frame is then taken as the start of a candidate repeated sequence. We then compare each frame in this candidate sequence with the corresponding frame at the beginning of the video; frames whose $L_2$ distance is below the threshold are counted as repetitions. Empirically, a threshold of 55 was found to align better with human perception and was consequently applied to both steps. Finally, we report the mean NoRepeat Score across all videos as the final result. The detailed implementation code is included in the supplementary material.

\paragraph{The implementation of RIFLEx and UltraViCo on Wan.}
Since Wan does not exhibit content repetition, it is not applicable to determine the dominant frequency from the repetition period as performed in \cite{zhao2025riflex}. Instead, following Sec.~\ref{sec:repetition_mechanism}, we take the largest-amplitude frequency $\phi_0$ as the dominant frequency.

For UltraViCo, the first frame's decay factor is set negative to fix its blurring. We hypothesize that this is caused by the causal design of the video VAE, where the first frame is encoded independently and without temporal compression. As a result, it exhibits different statistical properties from subsequent frames and becomes more sensitive to perturbations.

\paragraph{Details of the ablation study.} 
Herein, we detail the setup of the ablation study in Sec.~\ref{sec: experiments}. Specifically, as shown in Fig.~\ref{fig:ablation} (top), we compare three decay strategies—parabolic, linear, and constant. The parabolic strategy takes the following form:
\begin{equation}
S'_{ij} = \lambda_{ij} \cdot S_{ij}, \quad \text{where} \quad \lambda_{ij} =
\begin{cases}
1, \quad \text{if } |i-j| \le L/2 \text{ or } S_{ij} < 0, \\
\alpha_1(|i-j|/L^\prime)^2+\alpha_2(1-(|i-j|/L^\prime)^2),\quad \text{otherwise}, 
\end{cases}
\end{equation}
whereas the linear strategy takes the following form:
\begin{equation}
S'_{ij} = \lambda_{ij} \cdot S_{ij}, \quad \text{where} \quad \lambda_{ij} =
\begin{cases}
1, \quad \text{if } |i-j| \le L/2 \text{ or } S_{ij} < 0, \\
\alpha_1|i-j|/L^\prime+\alpha_2(1-|i-j|/L^\prime),\quad \text{otherwise}, 
\end{cases}
\end{equation}
and the constant strategy is 
\begin{equation}
S'_{ij} = \lambda_{ij} \cdot S_{ij}, \quad \text{where} \quad \lambda_{ij} =
\begin{cases}
1, \quad \text{if } |i-j| \le L/2 \text{ or } S_{ij} < 0, \\
\alpha,\quad \text{otherwise}.
\end{cases}
\end{equation}
We set $\alpha = 0.9$ for the constant strategy, and $\alpha_1=0.85, \alpha_2=0.95$ for the parabolic and the linear strategies. As shown in Fig.~\ref{fig:ablation} (top), parabolic, linear, and constant decay yield only minor differences, indicating that the key is distinguishing in-window from out-of-window tokens rather than the decay shape.

\subsection{Additional Experiments of Different Extrapolation Ratios and Models}
\label{appendix: more_quantitive_experiments}

\paragraph{Settings.} In this section, we provide some additional extrapolation ratios from $s=2$ to $5$ and models based on 25 prompts from VBench~\citep{huang2024vbench}. To evaluate the generality of UltraViCo, we test $2\times$ extrapolation on HunyuanVideo, Wan, and CogVideoX, as well as $3\times$ and $4\times$ extrapolation on CogVideoX. In addition, we assess $5\times$ extrapolation on HunyuanVideo. For Wan, we set $\alpha=0.9$. For HunyuanVideo, we use $\gamma=4$ across all ratios, with $\alpha=0.95,\beta=0.6$ at $2\times$ and $\alpha=0.9,\beta=0.8$ at $5\times$. For CogVideoX, we use $\gamma=1$ and $\beta=0.6$ for all ratios, with $\alpha=0.9$ at $2\times$ and $3\times$, and $\alpha=0.85$ at $4\times$. The configurations of other baselines follow Sec.~\ref{sec: setup}.

\paragraph{Results.} 
We compare UltraViCo with the baselines in Sec.~\ref{sec: experiments}. As shown in Tab.~\ref{tb: more_quantitive_experiments}, UltraViCo achieves the best performance across all models and extrapolation ratios, not only avoiding content repetition but also substantially improving video quality. For example, CogVideoX exhibits nearly static videos at 4× extrapolation (Dynamic Degree $\le$ 16) with poor visual quality (Imaging Quality $\le$ 56), whereas our method significantly enhances both temporal dynamics and visual quality, with Dynamic Degree and Imaging Quality improving by 200\% and 13.48\%, respectively. Furthermore, at $5\times$ extrapolation, UltraViCo also demonstrates strong performance, surpassing the best baseline scores by 350\% in Dynamic Degree and 47.59\% in Imaging Quality, indicating the potential of our method to extend to larger extrapolation ratios.
\begin{table}[h!]
  \centering
  \caption{\textbf{Quantitative results on VBench for more models and extrapolation}. 
  Note that NoRepeat Score is essentially a binary indicator: red entries indicate visually obvious repetitions, while others show no noticeable repetition. 
  }
  \label{tb: more_quantitive_experiments}
  \renewcommand\arraystretch{1.0}
  \resizebox{\linewidth}{!}{
  \begin{tabular}{l@{\hskip 1em}cccc@{\hskip 2em}cccc}
    \toprule
    \multirow{2}{*}{Method} 
     & \multicolumn{4}{c}{Wan with $2\times$ extrapolation}
     & \multicolumn{4}{c}{CogVideoX with $3\times$ extrapolation} \\
    \cmidrule(lr){2-5} \cmidrule(lr){6-9}
     & \makecell{NoRepeat$\uparrow$} & \makecell{Dynamic$\uparrow$} & \makecell{Quality$\uparrow$}  & \makecell{Overall$\uparrow$}
     & \makecell{NoRepeat$\uparrow$} & \makecell{Dynamic$\uparrow$} & \makecell{Quality$\uparrow$}  & \makecell{Overall$\uparrow$} \\
    \midrule
    PE & N/A & 32 & 58.13 & 23.22 & \cellcolor{red!10}82.52& 16 & 57.91 & 19.59 \\
    PI & N/A & 32 & 54.23 & 21.52 & 99.07 & 4 & 54.27 & 18.17 \\
    NTK & N/A & 44 & 59.59 & 23.52 & \cellcolor{red!10}86.07& 4 & 55.24 & 19.33 \\
    YaRN & N/A & 24 & 55.14 & 21.57 & 97.47 & 0 & 53.96 & 18.05 \\
    TASR & N/A & 36 & 59.97 & 23.70 & 97.93 & 8 & 55.75 & 19.24 \\
    RIFLEx & N/A & 16 & 48.15 & 20.34 & 97.86 & 8 & 55.31 & 19.03 \\
    \textbf{Ours} & N/A & \textbf{68} & \textbf{66.88} & \textbf{25.28}
                 & 99.38 & \textbf{32} & \textbf{60.09} & \textbf{24.77} \\
    \midrule
    % \addlinespace[0.6em]
    \multirow{2}{*}{Method} 
     & \multicolumn{4}{c}{HunyuanVideo with $2\times$ extrapolation}
     & \multicolumn{4}{c}{CogVideoX with $4\times$ extrapolation} \\
    \cmidrule(lr){2-5} \cmidrule(lr){6-9}
     & \makecell{NoRepeat$\uparrow$} & \makecell{Dynamic$\uparrow$} & \makecell{Quality$\uparrow$}  & \makecell{Overall$\uparrow$}
     & \makecell{NoRepeat$\uparrow$} & \makecell{Dynamic$\uparrow$} & \makecell{Quality$\uparrow$}  & \makecell{Overall$\uparrow$} \\
    \midrule
    PE & \cellcolor{red!10}80.43& 40 & 62.67 & 24.36 & \cellcolor{red!10}76.57& 16 & 55.25 & 17.27 \\
    PI & 98.87 & 4 & 52.35 & 23.55 & 88.53 & 4 & 46.82 & 16.63 \\
    NTK & 94.97 & 32 & 65.47 & 24.62 & \cellcolor{red!10}78.89& 2 & 52.74 & 18.14 \\
    YaRN & 97.99 & 4 & 52.87 & 23.26 & 94.75 & 4 & 47.36 & 16.90 \\
    TASR & 94.85 & 36 & 64.55 & 24.59 & 99.13 & 16 & 46.75 & 17.28 \\
    RIFLEx & 97.27 & 36 & 65.19 & 24.52 & 97.00 & 12 & 50.59 & 16.66 \\
    Ours & 97.53 & \textbf{44} & \textbf{66.50} & \textbf{24.82}
         & 96.79 & \textbf{48} & \textbf{62.70} & \textbf{25.39} \\
    \midrule
    % \addlinespace[0.6em]
    \multirow{2}{*}{Method} 
    & \multicolumn{4}{c}{CogVideoX with $2\times$ extrapolation}
     & \multicolumn{4}{c}{HunyuanVideo with $5\times$ extrapolation} \\
    \cmidrule(lr){2-5} \cmidrule(lr){6-9}
     & \makecell{NoRepeat$\uparrow$} & \makecell{Dynamic$\uparrow$} & \makecell{Quality$\uparrow$}  & \makecell{Overall$\uparrow$}
     & \makecell{NoRepeat$\uparrow$} & \makecell{Dynamic$\uparrow$} & \makecell{Quality$\uparrow$}  & \makecell{Overall$\uparrow$} \\
    \midrule
    PE & \cellcolor{red!10}92.31& 28 & 64.28 & 22.83 & \cellcolor{red!10}30.78& 4 & 39.04 & 15.64 \\
    PI & 98.85 & 8 & 57.11 & 21.88 & 81.58 & 0 & 36.63 & 16.76 \\
    NTK & \cellcolor{red!10}94.66 & 16 & 63.04 & 23.55 & 71.54 & 8 & 43.43 & 17.78 \\
    YaRN & 98.81 & 8 & 58.83 & 21.81 & 77.70 & 0 & 37.88 & 17.85 \\
    TASR & 95.91 & 16 & 62.17 & 23.44 & \cellcolor{red!10}35.31& 8 & 42.88 & 17.88 \\
    RIFLEx & 99.42 & 16 & 60.30 & 23.28 & 53.65 & 4 & 40.55 & 15.71 \\
    \textbf{Ours} & 98.92 & \textbf{32} & \textbf{64.39} & \textbf{25.36}
                 & 99.44 & \textbf{36} & \textbf{64.10} & \textbf{24.16} \\
    \bottomrule
  \end{tabular}
  }
\end{table}

\subsection{More Qualitative Results of Our Method}
\label{appendix: more_qualitive_experiments_wan_cog_3_4}
\begin{figure}
    \centering
    \includegraphics[width=\linewidth]{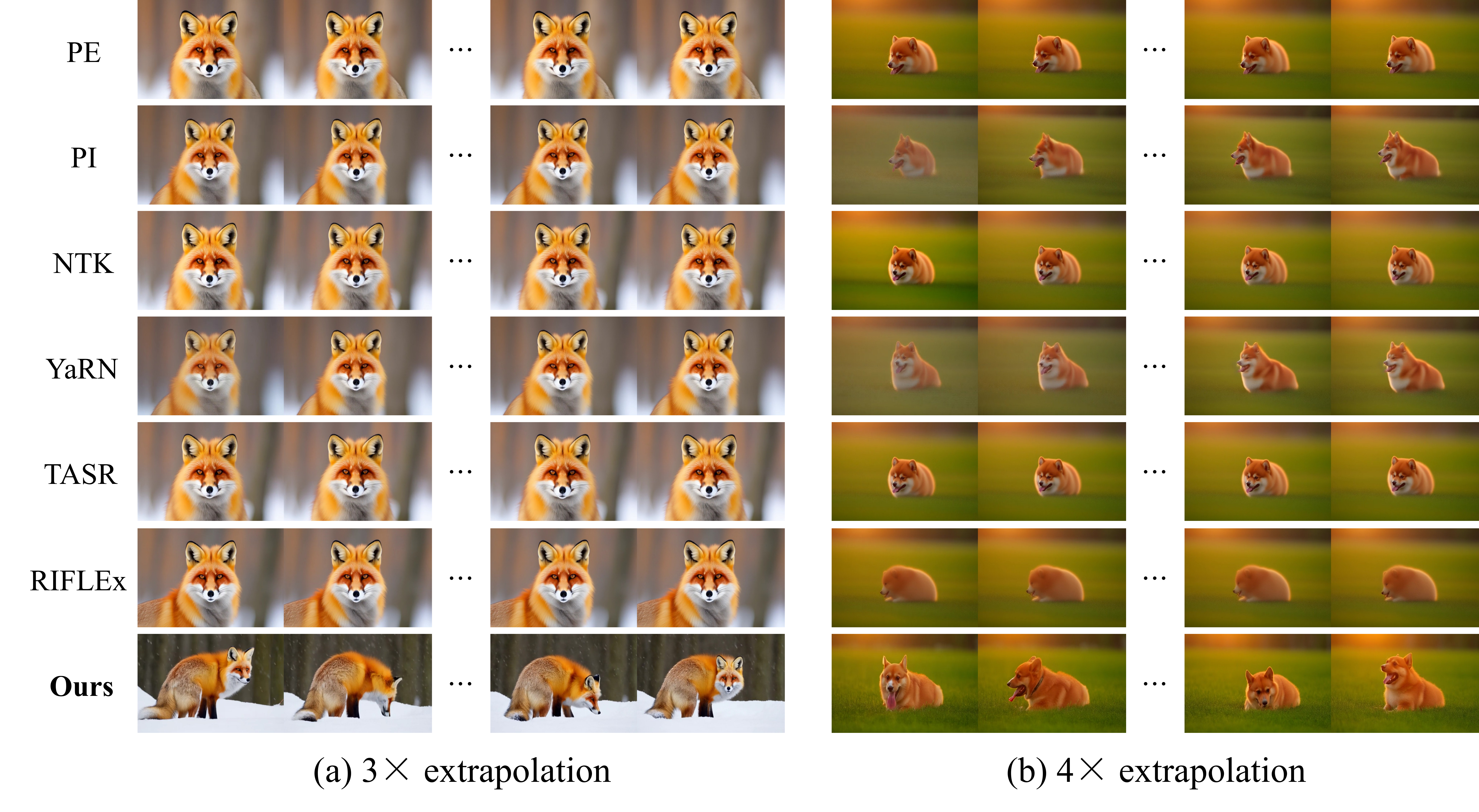}   
        \vspace{-.7cm}
        \caption{\textbf{Qualitative results on Wan}. The baselines produce nearly static videos with poor visual quality, whereas our method achieves significantly better quality and much more motion.}

    \label{fig: appendix-wan}
\end{figure}

\begin{figure}
    \centering
    \includegraphics[width=\linewidth]{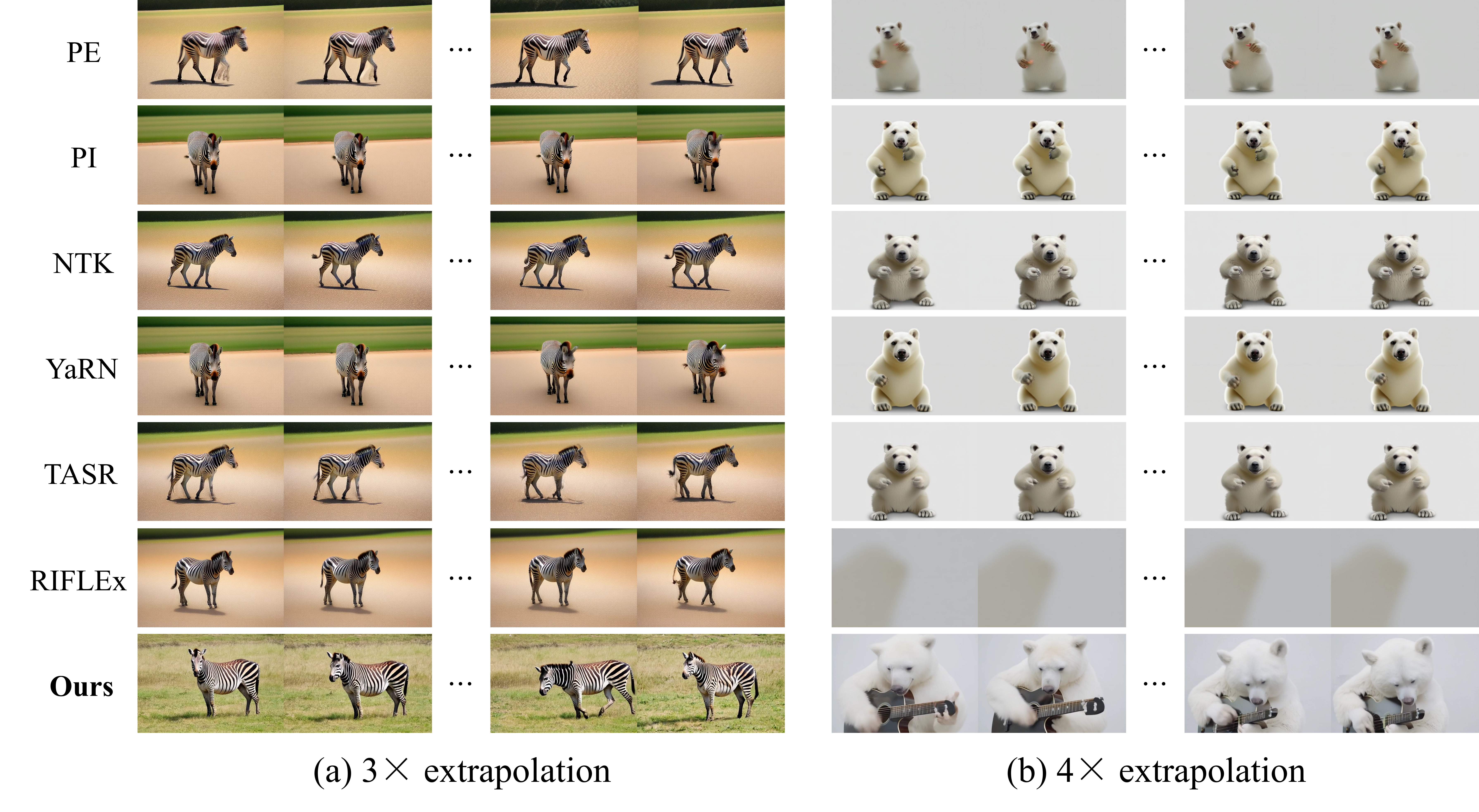}   
       
        \caption{\textbf{Qualitative results on CogVideoX}. The baselines produce nearly static videos with poor visual quality, whereas our method generates realistic results with rich details and fluid motion.}

    \label{fig: appendix-cogvideo}
\end{figure}

In this section, we provide additional qualitive results for the experiments in Sec.~\ref{sec: experiments}. As shown in Fig.~\ref{fig: appendix-wan} and Fig.~\ref{fig: appendix-cogvideo}, whether under $3\times$ or $4\times$ extrapolation ratios, and across Wan and CogVideoX, our method consistently achieves substantially superior visual quality and temporal dynamics compared to the baselines. For example, as shown in Fig.~\ref{fig: appendix-wan}, the videos generated by various baselines for 3× and 4× extrapolation on Wan are nearly completely static, whereas our method produces highly fluid and natural large-scale motion. Similarly, as shown in Fig.~\ref{fig: appendix-cogvideo}, the videos from the baselines are very blurry with dull colors, while our method generates realistic, natural results with rich details.

Moreover, we present another downstream task in Fig.~\ref{fig: controllable-pose}, where generation is performed based on a given pose. Our method achieves high quality and dynamic results while closely following the given conditions.

\subsection{Acceleration of UltraViCo via Sparse Attention and Distillation}

Building upon recent advances in sparse-attention-based video acceleration and distillation~\citep{fastvideo2024}, UltraViCo achieves about $16\times$ speed-up without compromising performance (see Table~\ref{tb:fastvideo_hunyuan}).

\subsection{Runtime and Memory Cost}
As shown in Table~\ref{tb:runtime_memory}, built on top of FlashAttention~\citep{dao2022flashattention} and SageAttention~\citep{zhang2024sageattention,zhang2024sageattention2}, UltraViCo incurs almost no additional overhead in either latency or memory usage.

\begin{figure}[t!]
    \centering
    \includegraphics[width=\linewidth]{images/controllable_pose.pdf}   
        \vspace{-.3cm}
        \caption{\textbf{Our method for pose-guided video generation}. Our method closely aligns with the given pose conditions, while ensuring high dynamic range and excellent visual quality.}

    \label{fig: controllable-pose}
\end{figure}

\section{Further details of UltraViCo}

\subsection{UltraViCo with Effieient Online Attention}
\label{appendix:algorithm}
UltraViCo does not require materializing the full attention matrix and can be seamlessly integrated into efficient online attention kernels. Herein, we present its implementation based on FlashAttention, as illustrated by Algorithm \ref{alg:online attn}.

\begin{algorithm}[h!]
\small
    \caption{\small UltraViCo FlashAttention Kernel}
    \label{alg:online attn}
    \begin{algorithmic}[1]
    \Require Matrices $Q, K, V \in \mathbb{R}^{N \times d}$, block size $b_q, b_{kv}$.
        \State Divide {$Q$} into $T_m = {N}/{b_q}$ blocks {$\{Q_m\}$}, and divide {$K$}, $V$ into $T_n = {N}/{b_{kv}}$ blocks {$\{K_n\}$} and $\{V_n\}$;
                
    \For {$\textbf{m}$ in [1, $T_m$]}
        \For {$\textbf{n}$ in [1, $T_n$]}

            \State $\vec i = m \times b_q + \text{range}(0, b_q), ~ \vec j = n \times b_{kv} + \text{range}(0, b_{kv}), ~~ \vec i \in \mathbb{R}^{1 \times b_q}, \vec j \in \mathbb{R}^{1 \times b_{kv}} $ ; 

            \State Initialize $ \lambda \in \mathbb{R}^{b_q \times b_{kv}}$ to 0 ; 

            \State $ \lambda = \text{Eq}.~\ref{eq:focus_attention_final}(\vec i, \vec j) $ ; 
        
            \State $S_m^n =\lambda Q_m K_n^T$ ;

            \State $p_m^n = \mathrm{max}(p_m^{n-1}, \mathrm{rowmax}(S_m^n))$ ;
            
            \State $ \widetilde P_m^n = \mathrm{exp}(S_m^n - p_m^n)$ ;

            \State $l_m^n = e^{p_m^{n-1}-p_m^n}\,l_m^{n-1} + \mathrm{rowsum}(\widetilde P_i^j)$ ;

            \State $O_m^n = \mathrm{diag}(e^{p_m^{n-1}-p_m^n})O_m^{n-1} + {\widetilde P_m^n} V_n$ ;
        \EndFor
        
        \State $O_m = \mathrm{diag}(l_m^{T_n})^{-1} O_m^{T_n}$ ;
         
    \EndFor
    
    \State \textbf{return} $O = \{O_m\}$;

    \end{algorithmic}
    
\end{algorithm}

\subsection{Ablation on hyperparameters}
\label{appendix:full_ablation}
\begin{figure}[h!]
  \centering

  \begin{subfigure}{\textwidth}
    \centering
    
      \centering
      \includegraphics[width=\linewidth,keepaspectratio]{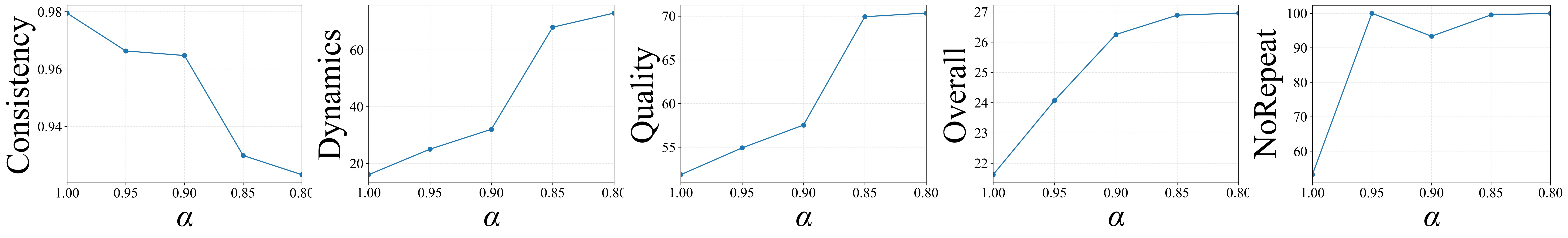}
    
    \subcaption{Schematic diagram of the  $\alpha$ sensitivity curve.}
  \end{subfigure}

  \begin{subfigure}{\textwidth}
    \centering
    
      \centering
      \includegraphics[width=\linewidth,keepaspectratio]{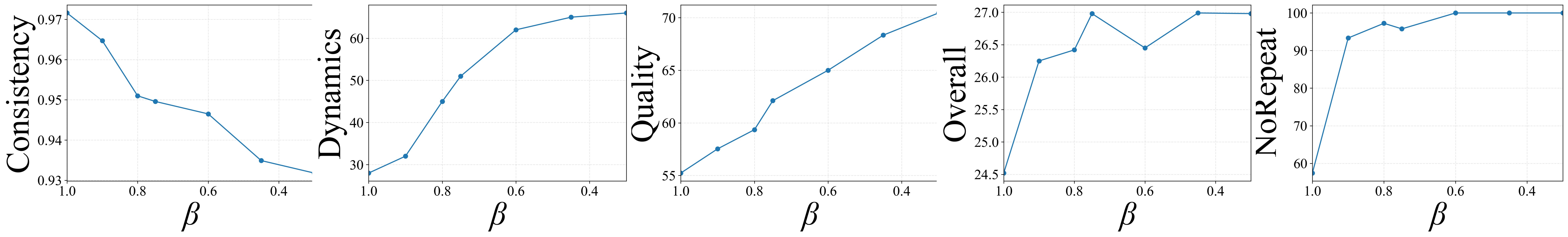}
  
    \subcaption{Schematic diagram of the  $\beta$ sensitivity curve.}
  \end{subfigure}

  \caption{\textbf{Illustration of the hyperparameter sensitivity curve.}}
  \label{fig:appendix-alphabet}
\end{figure}

In this section, we present more detailed illustrative ablation results for the hyperparameters $\alpha$ and $\beta$. The detailed sensitivity curve is shown in Fig. 
\ref{fig:appendix-alphabet}, while the illustrative ablations on the independent effects of $\alpha$ and $\beta$ in the main experiments are reported in Tab. \ref{tb:ablation_on_alphabet}.

\begin{table}[h!]
 \centering
 \caption{\textbf{Illustrative sensitivity analysis of $\alpha$ on Hunyuan at $3\times$ extrapolation. We set $\beta$ equal to $\alpha$, i.e., a single decay factor is shared globally.}}
 \label{tb:rebuttal_alpha_sensitivity}
 \renewcommand\arraystretch{1.0}
 \begin{tabular}{lccccc}
 \toprule
 \makecell{$\alpha$}
 & \makecell{Consistency$\uparrow$} 
 & \makecell{Dynamics$\uparrow$} 
 & \makecell{Quality$\uparrow$} 
 & \makecell{Overall$\uparrow$} 
 & \makecell{NoRepeat$\uparrow$} \\
 \midrule
 1.0  & 0.9795 & 16 & 51.85 & 21.62 & 53.17 \\
 0.95 & 0.9663 & 25 & 54.92 & 24.07 & 100   \\
 0.9  & 0.9647 & 32 & 57.53 & 26.25 & 93.34 \\
 0.85 & 0.9298 & 68 & 69.93 & 26.89 & 99.53 \\
 0.8  & 0.9231 & 73 & 70.35 & 26.96 & 100   \\
 \bottomrule
 \end{tabular}
\end{table}

\begin{table}[h!]
 \centering
 \caption{\textbf{Illustrative sensitivity analysis of $\beta$ on Hunyuan at $3\times$ extrapolation. We set $\alpha=0.9$ across all settings.}}
 \label{tb:rebuttal_beta_sensitivity}
 \renewcommand\arraystretch{1.0}
 \begin{tabular}{lccccc}
 \toprule
 \makecell{$\beta$}
 & \makecell{Consistency$\uparrow$} 
 & \makecell{Dynamics$\uparrow$} 
 & \makecell{Quality$\uparrow$} 
 & \makecell{Overall$\uparrow$} 
 & \makecell{NoRepeat$\uparrow$} \\
 \midrule
 1.0  & 0.9716 & 28 & 55.23 & 24.52 & 57.42 \\
 0.9  & 0.9647 & 32 & 57.53 & 26.25 & 93.34 \\
 0.8  & 0.9510 & 45 & 59.35 & 26.42 & 97.25 \\
 0.75 & 0.9496 & 51 & 62.11 & 26.98 & 95.77 \\
 0.6  & 0.9465 & 62 & 65.00 & 26.45 & 100   \\
 0.45 & 0.9349 & 65 & 68.34 & 26.99 & 100   \\
 0.3  & 0.9318 & 66 & 70.45 & 26.98 & 100   \\
 \bottomrule
 \end{tabular}
\end{table}

\begin{table}[h!]
 \centering
\caption{\textbf{Illustrative ablation experiments that independently examine the individual effects of $\alpha$ and $\beta$.}}
 \label{tb:ablation_on_alphabet}
 \renewcommand\arraystretch{1.0}
 \begin{tabular}{lccccc} 
 \toprule
\makecell{Method}
& \makecell{Consistency$\uparrow$} &  \makecell{Dynamics$\uparrow$} & \makecell{Quality$\uparrow$} 
& \makecell{Overall$\uparrow$} 
& \makecell{NoRepeat$\uparrow$} 
\\ 
\midrule
\multicolumn{6}{c}{HunyuanVideo with $3\times$ extrapolation } \\
\midrule
$\alpha=1,\beta=1$ &0.9795& 16 &	51.85& 21.62 & 53.17\\
$\alpha=0.9,\beta=1$ &0.9716& 28 &55.23& 24.52& 57.42\\
$\alpha=1,\beta=0.6$ &0.9784& 25	 &55.13& 23.13& 93.52\\
$\alpha=0.9,\beta=0.6$ &0.9465& 62	&65.00&26.45&100\\
\midrule
\multicolumn{6}{c}{Wan2.1-1.3B with $3\times$ extrapolation } \\
\midrule
$\alpha=1$ &0.9419&6&56.28&18.53 & --\\
$\alpha=0.9$ &0.9444& 46 &62.43& 23.21& --\\
\bottomrule
 \end{tabular}

\end{table}

\begin{table}[h!]
 \centering
 \caption{\textbf{Illustrative performance when combined with recent video-acceleration methods on HunyuanVideo.}}
 \label{tb:fastvideo_hunyuan}
 \renewcommand\arraystretch{1.0}
 \begin{tabular}{lcccccc}
 \toprule
 \makecell{Setting}
 & \makecell{Time Cost$\downarrow$}
 & \makecell{Consistency$\uparrow$}
 & \makecell{Dynamics$\uparrow$}
 & \makecell{Quality$\uparrow$}
 & \makecell{Overall$\uparrow$}
 & \makecell{NoRepeat$\uparrow$} \\
 \midrule
 3$\times$                          & 5 GPU$\cdot$hours   & 0.9465 & 62 & 65.00 & 26.45 & 100    \\
 3$\times$ with FastVideo          & 0.3 GPU$\cdot$hours & 0.9432 & 64 & 63.89 & 25.98 & 100    \\
 4$\times$                        & 8 GPU$\cdot$hours   & 0.9491 & 42 & 66.54 & 24.52 & 99.87  \\
 4$\times$ with FastVideo          & 0.5 GPU$\cdot$hours & 0.9399 & 40 & 62.24 & 24.83 & 96.32  \\
 \bottomrule
 \end{tabular}
\end{table}

\begin{table}[h!]
 \centering
 \caption{\textbf{Illustrative runtime and memory comparison.} Note that SageAttention is optimized for 4090-like architectures; on A800, its runtime is comparable to FlashAttention. }
 \label{tb:runtime_memory}
 \renewcommand\arraystretch{1.0}
 \begin{tabular}{lcc}
 \toprule
 \makecell{Model / Method} 
 & \makecell{Time (s / iter)} 
 & \makecell{Memory (per GPU)} \\
 \midrule
 \multicolumn{3}{c}{{HunyuanVideo ($3\times$ extrapolation)}} \\
 \midrule
 SageAttention              & 341.2 & 73188M \\
 SageAttention + Ours       & 349.6 & 72346M \\
 FlashAttention             & 349.3 & 76030M \\
 FlashAttention + Ours      & 355.3 & 75932M \\
 \midrule
 \multicolumn{3}{c}{{Wan ($3\times$ extrapolation)}} \\
 \midrule
 SageAttention              & 32.13 & 24342M \\
 SageAttention + Ours       & 34.12 & 24342M \\
 FlashAttention             & 32.64 & 24349M \\
 FlashAttention + Ours      & 33.74 & 24346M \\
 \bottomrule
 \end{tabular}
\end{table}

\begin{figure}[h!]
  \centering
\begin{subfigure}{0.9\textwidth}
    \centering
    
      \centering
      \includegraphics[width=\linewidth,keepaspectratio]{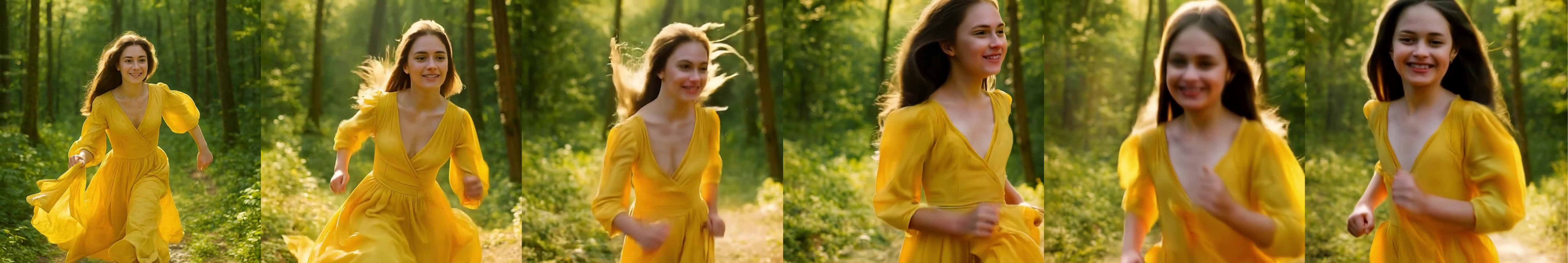}
  
    \subcaption{Performance of the video-continuation baseline alone.}
  \end{subfigure}
  
  \begin{subfigure}{0.9\textwidth}
    \centering
    
      \centering
      \includegraphics[width=\linewidth,keepaspectratio]{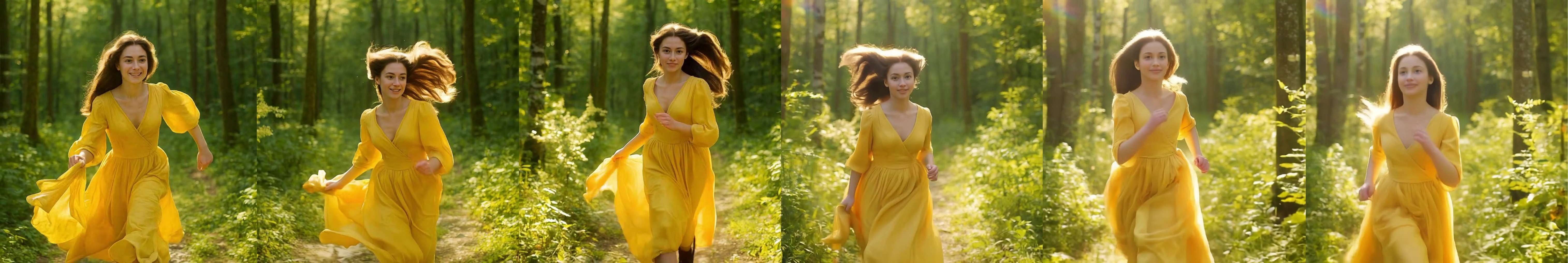}
    
    \subcaption{Illustration of combining UltraViCo with the video-continuation method.}
  \end{subfigure}

  % \vspace{-0.5cm}

  \caption{\textbf{Application of UltraViCo to segment-wise long-video generation.} (a) Wan2.2-TI2V uses only a few ending frames, causing identity drift; (b) UltraViCo alleviates this issue.}
  \label{fig:wan2.2}

\end{figure}

\end{document}